\documentclass[sigconf]{acmart}
\usepackage{xcolor}

\usepackage{amsmath}

\usepackage{amssymb}
\usepackage{pifont}
\usepackage{multirow}
\usepackage{threeparttable}
\usepackage{booktabs}
\usepackage{algorithm}
\usepackage{algorithmic}
\usepackage{listings}
\usepackage{tabularx}
\usepackage{array}
\usepackage{booktabs}
\usepackage{fontawesome5}

\setcopyright{none}
\renewcommand{\footnotetextcopyrightpermission}[1]{}
\acmDOI{}
\acmISBN{}
\acmYear{}
\copyrightyear{}
\acmConference[]{}{}{}

\AtBeginDocument{%
  \fancypagestyle{standardpagestyle}{\fancyhf{}}%
  \fancypagestyle{firstpagestyle}{\fancyhf{}}%
  \pagestyle{empty}%
}

\title{Normative Deviation Modeling for EEG-based \\ Depression Detection}

\author{\small Wan Jiang$^{1}$, Jingqi Cheng$^{2}$, Hui Zhang$^{1}$, Xu Zhao$^{2}$, Xiaojing Chen$^{2}$, Shengeng Tang$^{1}$, Jingjing Wu$^{1}$, Richang Hong$^{1}$}
\affiliation{%
  \institution{\small $^{1}$Hefei University of Technology \qquad $^{2}$Anhui University}\country{}
}

\begin{document}
\begin{abstract}
Web-enabled digital health services are expanding access to physiological monitoring and mental health screening. However, electroencephalography (EEG)-based major depressive disorder (MDD) detection remains challenging across heterogeneous data sources, where individual variability and acquisition differences may affect disease-related representations. In this work, we investigate the use of healthy normative references for EEG-based MDD detection. Specifically, structural and distributional reference models are independently trained using healthy control (HC) recordings to characterize departures from healthy EEG patterns. We develop two diagnostic approaches that incorporate normative deviations at different representation levels. Scalar-level Normative Deviation-guided Diagnosis (S-NDD) summarizes structural and distributional deviations into scalar scores and integrates them with deep EEG representations. Feature-level Normative Deviation-guided Diagnosis (F-NDD) retains dimension-wise residuals and combines them with deep EEG representations through attention-based aggregation and auxiliary supervision. We also employ target HC-assisted score calibration for cross-source evaluation without updating the trained models. Experiments on two public EEG datasets examine both approaches under subject-independent within-source and bidirectional cross-source settings. The results demonstrate the effectiveness of
normative deviation-guided diagnosis across
different evaluation settings.
\end{abstract}
\keywords{EEG-based Depression Detection, Healthy Normative Modeling, Normative Deviation, Cross-source Diagnosis, Digital Health}
\maketitle
\raggedbottom
\section{Introduction}
\label{sec:intro}

\begin{figure}[t]
    \centering
\includegraphics[width=\linewidth]{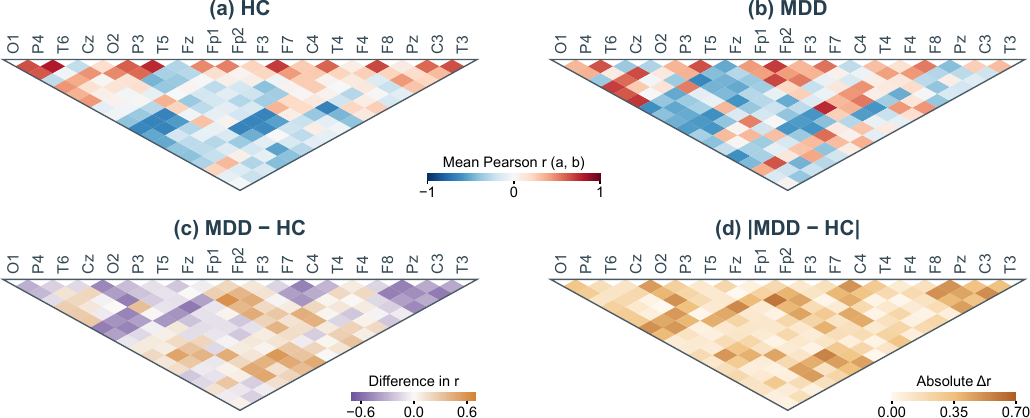 }
\setlength{\abovecaptionskip}{-5pt} 
    \caption{Inter-electrode correlation patterns
in HC and MDD EEG recordings.
(a) and (b) show the mean Pearson correlation
matrices for HC and MDD, respectively.
(c) and (d) show their signed and absolute
differences (MDD$-$HC), respectively.
All panels use the same electrode ordering
derived from HC-based hierarchical clustering.}
    \label{fig:moti}
    \vspace{-1.5em}
\end{figure}

Web-enabled digital health services have created new opportunities for disease screening and health monitoring beyond conventional clinical settings~\cite{wang2026sustainable,fu2026towards}. The demand for such services is particularly urgent in Major Depressive Disorder (MDD) screening, given the limited access to mental healthcare and persistent stigma-related barriers~\cite{globaldisability,who2025mentalhealth}. Among various screening modalities, electroencephalography (EEG)-based MDD detection has attracted considerable attention, as EEG provides objective physiological measurements and generates digital signals that can be readily transmitted for remote analysis~\cite{rehman2025advances,EEG}. Recent deep learning methods have achieved promising results by modeling the temporal, spectral, and spatial characteristics of EEG signals~\cite{xu2026major,khayretdinova2024prediction,choudhary2024spatio}. Despite these advances, ensuring reliable EEG-based MDD detection across diverse subjects and recording conditions remains a significant challenge.

This challenge primarily arises from the inherent characteristics of EEG signals, which are non-stationary, noisy, and subject-dependent. Disease-related neural patterns are often intertwined with individual physiological variability and acquisition artifacts~\cite{EEGSignal}. Consequently, supervised models may capture subject-specific or recording-related patterns alongside disease-related information, limiting their ability to generalize to unseen subjects and independently collected datasets~\cite{lawal2026transfer}. The limited number of subjects in psychiatric EEG datasets, coupled with the high cost of clinical data collection and annotation, further exacerbates these generalization difficulties. These limitations highlight the need for learning disease-discriminative representations that remain robust to subject-specific and acquisition-related variations.

A natural way to mitigate this data limitation is to enlarge the training set through data augmentation. In particular, generative data augmentation has attracted increasing attention in EEG analysis, where generative models synthesize additional EEG samples for training~\cite{wang2024diffmdd,zhao2024eeg}. However, generating realistic EEG signals remains challenging, especially in low-data settings~\cite{you2025virtual}. Synthetic EEG samples may exhibit artifacts and deviations from the statistical properties of authentic recordings, raising concerns about their fidelity~\cite{habashi2023generative,choi2026fail}. Moreover, the benefits of synthetic augmentation for downstream MDD classification may vary across datasets~\cite{carrle2023generation}. These concerns suggest that increasing data quantity alone does not necessarily ensure more reliable EEG-based MDD detection.

Clinical studies have reported differences in EEG spectral activity and functional connectivity between individuals with MDD and healthy controls (HC)~\cite{leuchter2012resting,das2020resting}. Consistent with these findings, our preliminary analysis reveals differences in inter-electrode correlation patterns between the two groups, as shown in~\autoref{fig:moti}. These observations indicate that healthy and depressive EEG recordings exhibit distinct patterns of neural activity. Rather than treating HC recordings solely as a comparison group, their physiological patterns may also provide a reference for characterizing individual variations. Healthy normative modeling offers a means of establishing such references from HC recordings and quantifying deviations from them. These deviations may provide complementary information for identifying MDD-related abnormalities. Nevertheless, normative deviations do not necessarily translate into discriminative evidence for MDD detection. Their diagnostic utility depends on how disease-relevant variations are represented and integrated with EEG features, particularly across heterogeneous data sources.

To this end, we investigate normative
deviations at different levels of granularity,
each capturing distinct aspects of departures
from healthy EEG patterns. We therefore
develop two diagnostic formulations to examine
their respective contributions to MDD detection.
Both establish structural and distributional
references exclusively from healthy EEG
recordings. Specifically, we propose \textit{Scalar-level
Normative Deviation-guided Diagnosis} (S-NDD),
which summarizes structural and distributional
deviations into compact scores and integrates
them with deep EEG representations.
We further introduce \textit{Feature-level
Normative Deviation-guided Diagnosis} (F-NDD),
which preserves dimension-wise residuals
and incorporates them into deep EEG
representations through attention-based
aggregation and auxiliary supervision.
To accommodate cross-source shifts in diagnostic scores, both approaches support target HC-assisted score calibration without updating model parameters. We evaluate their diagnostic performance under within-source and bidirectional cross-source settings.

In summary, our key contributions are as follows:
\begin{itemize}
    \item We establish structural
    and distributional healthy references to
    characterize normative deviations in EEG
    recordings. These references are learned
    exclusively from HC and
    provide complementary information for
    MDD diagnosis.

    \item We introduce two normative deviation-guided
    diagnostic approaches, S-NDD and F-NDD.
    S-NDD integrates scalar deviation scores
    with deep EEG representations, while F-NDD
    incorporates dimension-wise residuals through
    shared attention aggregation and auxiliary
    diagnostic supervision.

    \item We conduct comprehensive evaluations
    on two public EEG datasets under
    subject-independent within-source and
    bidirectional cross-source settings.
    We further investigate target HC-assisted
    score calibration to accommodate
    cross-source variations in diagnostic scores.
\end{itemize}

\begin{figure*}[t]
  \centering
  \includegraphics[width=0.92\linewidth]{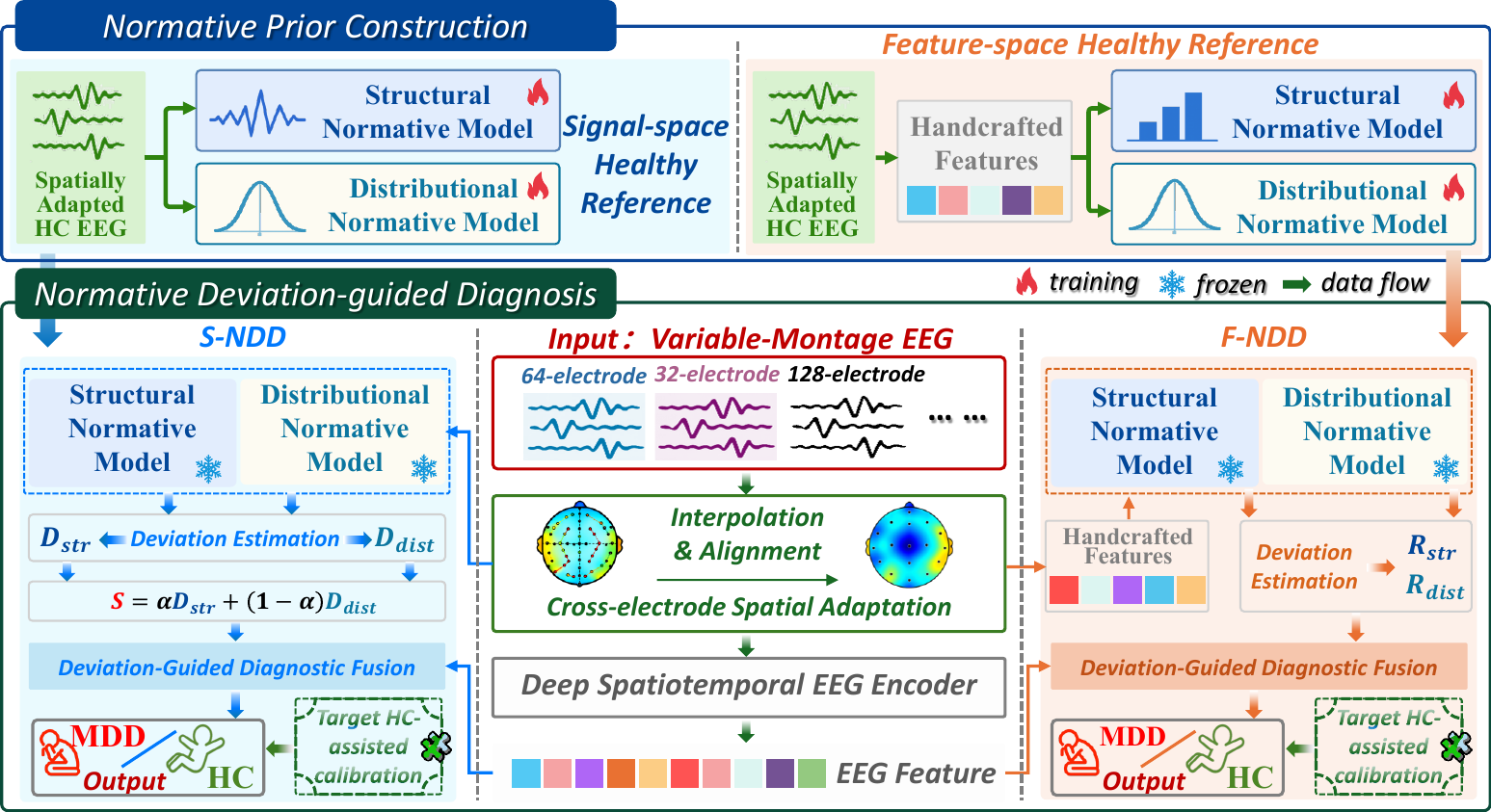 }
  \caption{Overview of S-NDD and F-NDD.
Both methods establish structural and distributional
healthy references in distinct representation spaces.
S-NDD quantifies scalar normative deviations for
deep EEG feature fusion, whereas F-NDD preserves
feature-level residuals for deviation-guided
diagnosis. Optional target HC-assisted calibration adjusts cross-dataset predictions.}
 \vspace{-1em}
  \label{fig:framework}
\end{figure*}

\section{Related Work}
\label{sec:related_work}

\subsection{EEG-based MDD Detection}
EEG-based MDD detection has progressed from
handcrafted features to deep representation
learning. Early CNN-based methods learned
discriminative EEG features~\cite{acharya2018depression},
while graph-based approaches, including
MAST-GCN~\cite{mastgcn_mdd}, adaptive temporal
graph networks~\cite{luo2023adaptive}, and
ST-GCA~\cite{11369854}, model spatial and
temporal dependencies. Hybrid architectures
combine graph convolution and Transformers,
as in SLiTRANet~\cite{singh2024slitranet},
or adopt lightweight convolutional
Transformers~\cite{hou2025lightweight}.
Temporal-spatial-spectral learning has also
been explored using emotion-elicited
EEG~\cite{cnn_transformer_mdd_2025}.
Research has further examined diverse acquisition
and clinical settings, including single-channel
detection~\cite{vaniya2026simplifying} and
multi-dataset diagnosis and treatment-response
prediction~\cite{olbrich2026deep}.
General EEG representation models, such as
EEG Conformer~\cite{song2023eegconformer}
and CBraMod~\cite{wang2025cbramod},
have advanced convolutional attention and
large-scale pretraining. Nevertheless,
discriminative EEG representations may retain
subject-specific and acquisition-related
variations, limiting generalization across
independently collected datasets.

\subsection{Normative Modeling and Healthy References}
Normative modeling characterizes individual observations relative to a reference population, providing a framework for investigating heterogeneity beyond average case--control differences~\cite{marquand2019conceptualizing,rutherford2022normative}. In psychiatric research, normative models characterize individual deviations from expected physiological or behavioral patterns~\cite{marquand2019conceptualizing}. Practical frameworks have also been developed to estimate and evaluate such deviations~\cite{rutherford2022normative}. Generative approaches further learn healthy data distributions and identify departures from these distributions for anomaly detection. For example, diffusion-based models have been trained on healthy brain images to detect and segment abnormalities in CT and MRI data~\cite{pinaya2022fast}.
These studies illustrate the potential of reference-based analysis for characterizing individual abnormalities. However, findings from neuroimaging do not directly establish the diagnostic utility of normative deviations in EEG-based MDD detection. Moreover, the representation and utilization of normative deviations may influence their effectiveness across different subjects and data sources.

\subsection{Cross-domain EEG Diagnosis}
Domain adaptation aims to reduce distribution
discrepancies between source and target domains.
Representative methods include DANN~\cite{ganin2016dann},
which learns domain-adversarial representations,
DAN~\cite{long2015dan}, which aligns distributions
through maximum mean discrepancy, and Deep
CORAL~\cite{sun2016deepcoral}, which matches
feature covariance statistics.
Cross-domain learning has also been explored
for EEG-based depression detection.
TFST-SDKA~\cite{xu2026tfstsdka} combines
temporal-frequency-spatial representations
with sub-domain knowledge alignment for
cross-subject detection. While cross-subject
evaluation examines unseen participants,
cross-dataset evaluation additionally involves
differences in acquisition protocols,
electrode configurations, and participant
populations. Evaluation protocols 
depend on the availability of target-domain
data for adaptation or calibration.
However, reducing distribution discrepancies
does not necessarily separate disease-related
variations from source-specific characteristics.
\section{Preliminaries}
\label{sec:Pre}
We consider EEG-based MDD detection in the context of Web-enabled digital health services. The objective is to identify MDD status by analyzing the brain activity of individual subjects. Given a labeled
source dataset
$\mathcal{D}_s=\{(X_i^s,y_i^s)\}_{i=1}^{N_s}$,
where $X_i^s\in\mathbb{R}^{C_s\times T_i}$
denotes the EEG recording of subject $i$,
$C_s$ and $T_i$ represent the number of
electrodes and recording length, respectively,
and $y_i^s\in\{0,1\}$ indicates HC or MDD
status. Within-source MDD detection involves training
and evaluating diagnostic models on disjoint
subjects from the same data source.
We further consider cross-source detection,
where a Web-based diagnostic service receives
EEG recordings from a new data source. Let $\mathcal{D}_t=\{X_j^t\}_{j=1}^{N_t}$
denote an independently collected target
dataset, where
$X_j^t\in\mathbb{R}^{C_t\times T_j}$
represents the recording of target subject $j$.
The source-trained diagnostic models are applied
to target subjects despite potential differences
in electrode configurations and acquisition
characteristics. Cross-source detection aims to achieve
reliable MDD diagnosis across heterogeneous
EEG data sources.

\section{Normative Deviation-guided Diagnosis}
Our framework establishes a normative reference from healthy EEG recordings to characterize individual deviations from healthy patterns. Rather than relying solely on supervised EEG representations, we incorporate structural and distributional deviations as complementary diagnostic information. As illustrated in ~\autoref{fig:framework}, we develop two formulations that represent normative deviations at different levels of granularity. S-NDD summarizes deviations into scalar scores, while F-NDD preserves fine-grained deviation features. Both formulations integrate the resulting deviations with learned EEG representations for MDD detection.

\subsection{Dual Healthy Reference Modeling}
We construct normative references exclusively
from healthy control (HC) recordings to
characterize departures from healthy EEG
patterns. Since a single reference model may
not capture all aspects of healthy variability,
we establish complementary structural and
distributional references. Specifically,
VQ-VAE~\cite{van2017neural,yang2025thd}
captures recurring patterns through discrete
latent quantization, while a diffusion
model~\cite{ho2020denoising,pinaya2022fast}
learns healthy data distributions through
noise-conditioned denoising. To accommodate
heterogeneous electrode configurations,
EEG recordings are first aligned to a common
spatial layout. S-NDD trains its normative
models on the aligned EEG signals, whereas
F-NDD independently trains corresponding models
on handcrafted features described in Section~\ref{subsec:feature}.

\noindent\textbf{Cross-electrode Spatial Adaptation.}
To accommodate heterogeneous electrode
configurations, we employ spherical spline
interpolation~\cite{klem1999ten} to align EEG
recordings with the standard 19-electrode
International 10--20 system. Given an EEG
recording $\mathbf{x}_{orig}\in
\mathbb{R}^{C_{orig}\times T}$, the adapted
signal $\mathbf{x}\in\mathbb{R}^{19\times T}$
provides a common electrode configuration
for normative modeling and diagnosis.
Implementation details are provided in
Appendix~\ref{app:spatial}.


\noindent\textbf{Structural Normative Modeling.}
We employ VQ-VAE~\cite{van2017neural,yang2025thd}
to capture recurring patterns in healthy EEG
representations. Given an HC input $\mathbf{u}$,
the encoder produces a latent representation
$\mathbf{z}_e$, which is quantized into
$\mathbf{z}_q$ through a learnable codebook.
The model is optimized using reconstruction,
codebook, and commitment losses:
\begin{equation}
\mathcal{L}_{str}
=
\operatorname{MSE}(\mathbf{u},D(\mathbf{z}_q))
+
\operatorname{MSE}(\mathrm{sg}[\mathbf{z}_e],\mathbf{z}_q)
+
\beta\operatorname{MSE}(\mathbf{z}_e,\mathrm{sg}[\mathbf{z}_q]).
\end{equation}
where $\operatorname{MSE}(\cdot,\cdot)$ denotes
the mean squared error, $D(\cdot)$ is the decoder,
$\mathrm{sg}[\cdot]$ denotes the stop-gradient
operator, and $\beta$ weights the commitment loss.
The trained VQ-VAE serves as the structural
normative reference for subsequent deviation
estimation.

\noindent\textbf{Distributional Normative Modeling.}
We employ a diffusion model~\cite{ho2020denoising,pinaya2022fast}
to characterize the distribution of healthy EEG
representations. Given an HC input $\mathbf{u}_0$,
the forward diffusion process generates
$\mathbf{u}_t=\sqrt{\bar{\alpha}_t}\mathbf{u}_0+
\sqrt{1-\bar{\alpha}_t}\boldsymbol{\epsilon}$,
where $\boldsymbol{\epsilon}\sim\mathcal{N}(0,I)$
and $\bar{\alpha}_t$ is the cumulative product
of noise retention coefficients. The model is trained by minimizing
\begin{equation}
\mathcal{L}_{\mathrm{dist}}=
\mathbb{E}_{t,\mathbf{u}_0,\boldsymbol{\epsilon}}
\left[
\operatorname{MSE}\left(
\boldsymbol{\epsilon},
\epsilon_\theta(\mathbf{u}_t,t)
\right)
\right].
\end{equation}
The trained diffusion model serves as the
distributional normative reference for
subsequent deviation estimation.


\subsection{Scalar-level Normative Deviation-guided Diagnosis (S-NDD)}
\label{subsec:sndd}

S-NDD characterizes individual EEG recordings through
scalar deviations from healthy normative references.
Given an EEG recording
$\mathbf{x}\in\mathbb{R}^{19\times T}$,
the structural deviation is measured by the
VQ-VAE reconstruction error:
\begin{equation}
D_{str}
=
\operatorname{MSE}
\left(\mathbf{x},D(Q(E(\mathbf{x})))\right),
\end{equation}
where $E(\cdot)$, $Q(\cdot)$, and $D(\cdot)$ denote
the encoder, quantizer, and decoder, respectively.
The distributional deviation
is measured by the diffusion denoising
reconstruction error:
\begin{equation}
D_{\mathrm{dist}} =
\operatorname{MSE}\left(
\mathbf{x},
\frac{
\mathbf{x}_t-\sqrt{1-\bar{\alpha}_t}
\epsilon_\theta(\mathbf{x}_t,t)
}{
\sqrt{\bar{\alpha}_t}+10^{-8}
}
\right),
\end{equation}
where $\mathbf{x}_t$ is the noisy EEG recording
obtained through the forward diffusion process,
and $\epsilon_\theta$ denotes the pretrained
noise prediction network.
The structural and distributional deviations are
combined into a scalar score:
$
S=\alpha D_{str}+(1-\alpha)D_{dist},
$
where $\alpha$ balances their contributions.
The score is normalized as
$S'=\operatorname{clip}\left(
\frac{S-P_5}{P_{95}-P_5},0,1
\right)$,
where $P_5$ and $P_{95}$ are the 5th and 95th
percentiles of training deviation scores.
A pretrained deep spatiotemporal EEG encoder
extracts features $\mathbf{F}_{\mathrm{deep}}$
from the spatially adapted EEG signal $\mathbf{x}$.
Deviation-guided fusion integrates these
features with the normalized deviation score
$S'$ as
$\mathbf{F}_S=\sigma(W_h\mathbf{F}_{\mathrm{deep}}
+W_sS'+\mathbf{b})$,
where $W_h$, $W_s$, and $\mathbf{b}$ are
learnable parameters. The fusion layer and diagnostic classifier
$C_{S,\phi}(\cdot)$ are jointly optimized
using cross-entropy loss:
\begin{equation}
\mathcal{L}_{\mathrm{diag}}^{S}
=
\operatorname{CE}(C_{S,\phi}(\mathbf{F}_S),y),
\end{equation}
where $\operatorname{CE}(\cdot,\cdot)$ denotes
the cross-entropy loss, $\phi$ represents
the classifier parameters, and $y$ is the
ground-truth label.
The pretrained EEG encoder and normative
models remain frozen.

\subsection{Feature-level Normative Deviation-guided Diagnosis (F-NDD)}
\label{subsec:feature}

While S-NDD summarizes departures from healthy
references using scalar deviation scores,
F-NDD characterizes deviations at the
individual EEG feature level. It preserves
dimension-wise structural and distributional
residuals for subsequent diagnostic learning.

\noindent\textbf{Handcrafted Feature Extraction.}
Given a spatially adapted EEG recording
$\mathbf{x}\in\mathbb{R}^{19\times T}$,
we extract spectral and inter-electrode
correlation features to characterize EEG
activity. Spectral features describe the
distribution of signal power across frequency
bands, while correlation features capture
statistical relationships between electrodes.
These features are concatenated into
$\mathbf{h}=H(\mathbf{x})\in\mathbb{R}^{456}$,
where $H(\cdot)$ denotes a deterministic
feature extractor. Detailed feature
definitions and extraction settings are
provided in Appendix~\ref{app:fndd_inference}. 

\noindent\textbf{Feature-level Deviation Modeling.}
Based on the extracted EEG features $\mathbf{h}$,
F-NDD estimates structural and distributional
deviations using two normative models independently
trained on source-domain HC features.
The structural residual is computed as
\begin{equation}
\mathbf{R}_{\mathrm{str}}
=
\mathbf{h}-D(Q(E(\mathbf{h}))).
\end{equation}
The distributional residual is estimated from
diffusion-based denoised features:
\begin{equation}
\mathbf{R}_{\mathrm{dist}}
=
\frac{1}{|\mathcal{T}|K}
\sum_{t\in\mathcal{T}}
\sum_{k=1}^{K}
\left(
\mathbf{h}-\widehat{\mathbf{h}}_0^{(t,k)}
\right),
\end{equation}
where $\mathcal{T}$ denotes the selected
diffusion timesteps, $K$ is the number of
noise realizations per timestep, and
$\widehat{\mathbf{h}}_0^{(t,k)}$ denotes the
denoised feature estimate at timestep $t$
under the $k$-th noise realization.
The structural and distributional residuals
are concatenated into a 912-dimensional
deviation representation:
$
\mathbf{F}_{\mathrm{dev}}
=
[\mathbf{R}_{\mathrm{str}};
\mathbf{R}_{\mathrm{dist}}]
\in\mathbb{R}^{912}.
$
The resulting representation is normalized
feature-wise using statistics estimated
exclusively from source-domain HC training data.

\noindent\textbf{Deviation-guided Diagnosis.}
F-NDD integrates deep EEG representations with
normative deviations through separate encoding
branches. Given the $i$-th spatially adapted
EEG window $\mathbf{x}_i$, a deep spatiotemporal
EEG encoder produces
$\mathbf{u}_i=E_{\mathrm{EEG}}(\mathbf{x}_i)$.
The structural and distributional residuals
are separately encoded and projected into
a unified deviation representation:
\begin{equation}
\mathbf{v}_i
=
P_{\mathrm{dev}}\left(
[E_{\mathrm{str}}(\mathbf{R}_{\mathrm{str},i});
E_{\mathrm{dist}}(\mathbf{R}_{\mathrm{dist},i})]
\right),
\end{equation}
where $E_{\mathrm{str}}$ and $E_{\mathrm{dist}}$
denote the respective residual encoders,
and $P_{\mathrm{dev}}$ is a learnable projection.
To maintain consistent window contributions
across the two representations, shared attention
weights are derived from the deep EEG features:
\begin{equation}
a_i=
\frac{\exp(\mathbf{w}^{\top}\mathbf{u}_i)}
{\sum_{j=1}^{N}\exp(\mathbf{w}^{\top}\mathbf{u}_j)},
\end{equation}
where $N$ denotes the number of windows
and $\mathbf{w}$ is a learnable parameter.
The resulting representations are
\begin{equation}
\mathbf{h}_{\mathrm{EEG}}
=\sum_{i=1}^{N}a_i\mathbf{u}_i,
\qquad
\mathbf{h}_{\mathrm{dev}}
=\sum_{i=1}^{N}a_i\mathbf{v}_i.
\end{equation}
They are concatenated as
$\mathbf{F}_F=
[\mathbf{h}_{\mathrm{EEG}};
\mathbf{h}_{\mathrm{dev}}]$,
which supports window-level prediction
with $N=1$ and subject-level prediction
through multi-window aggregation.
To encourage the deviation representation
to retain diagnostic information, an auxiliary
classification objective is imposed alongside
the fusion classifier:
\begin{equation}
\mathcal{L}_{\mathrm{diag}}^{F}
=
\operatorname{CE}(C_F(\mathbf{F}_F),y)
+
\lambda_{\mathrm{dev}}
\operatorname{CE}(C_{\mathrm{dev}}
(\mathbf{h}_{\mathrm{dev}}),y),
\end{equation}
where $C_F$ and $C_{\mathrm{dev}}$
denote the fusion and auxiliary classifiers,
respectively, and $\lambda_{\mathrm{dev}}$
controls the auxiliary contribution.
The normative models remain frozen during
diagnostic optimization.

\begin{table*}[t]
\centering
\caption{Performance comparison (\%) on Mumtaz2016
under subject-independent 10-fold cross-validation.
All baselines are evaluated using a unified protocol.
Results are mean $\pm$ std., with best and second-best
in bold and \underline{underlined}, respectively.}
\vspace{-0.5em}
\label{tab:sota_comparison}

\resizebox{\textwidth}{!}{%
\begin{tabular}{l|ccccc|ccccc}
\toprule
\multirow{2}{*}{\textbf{Methods}}
& \multicolumn{5}{c|}{\textbf{Window-level Metrics (\%)}}
& \multicolumn{5}{c}{\textbf{Subject-level Metrics (\%)}} \\
\cmidrule(lr){2-6} \cmidrule(lr){7-11}
& \textbf{Acc ($\uparrow$)}
& \textbf{Pre ($\uparrow$)}
& \textbf{Rec ($\uparrow$)}
& \textbf{F1 ($\uparrow$)}
& \textbf{AUC ($\uparrow$)}
& \textbf{Acc ($\uparrow$)}
& \textbf{Pre ($\uparrow$)}
& \textbf{Rec ($\uparrow$)}
& \textbf{F1 ($\uparrow$)}
& \textbf{AUC ($\uparrow$)} \\
\midrule

\multicolumn{11}{@{}c}{\textit{Discriminative Learning Methods}} \\
\hline

XGBoost \cite{hassan2024impact}
& $83.25 \pm 3.06$ & $83.54 \pm 7.49$
& $82.41 \pm 5.15$ & $82.49 \pm 5.85$
& $88.73 \pm 3.50$
& $89.17 \pm 4.66$ & $93.71 \pm 7.30$
& $86.33 \pm 5.53$ & $89.03 \pm 4.51$
& $92.19 \pm 3.92$ \\

1D-CNN-LSTM \cite{ay2019automated}
& $90.53 \pm 3.25$ & $90.17 \pm 6.48$
& $91.58 \pm 7.27$ & $90.52 \pm 3.72$
& $93.13 \pm 3.38$
& $93.26 \pm 5.30$ & $93.25 \pm 9.92$
& $95.83 \pm 6.80$ & $94.01 \pm 5.18$
& $95.13 \pm 3.43$ \\

1D-CNN-Trans \cite{cnn_transformer_mdd_2025}
& $85.56 \pm 4.61$ & $84.90 \pm 6.35$
& $87.97 \pm 5.76$ & $84.98 \pm 4.36$
& $92.39 \pm 5.03$
& $86.52 \pm 6.18$ & $86.02 \pm 6.69$
& $89.63 \pm 5.38$ & $86.21 \pm 5.24$
& $93.96 \pm 3.86$ \\

CWT-1D-CNN \cite{raghavendra2023automated}
& $91.54 \pm 5.36$ & $91.95 \pm 9.66$
& $92.30 \pm 8.40$ & $92.12 \pm 5.40$
& $92.57 \pm 5.41$
& $93.18 \pm 6.62$ & $92.80 \pm 10.20$
& $95.00 \pm 8.74$ & $93.31 \pm 6.47$
& $95.42 \pm 5.15$ \\

CWT-2D-CNN \cite{raghavendra2023automated}
& $90.50 \pm 4.61$ & $89.17 \pm 8.65$
& $93.63 \pm 6.25$ & $90.87 \pm 4.21$
& $94.49 \pm 2.08$
& $94.09 \pm 6.90$ & $94.46 \pm 9.83$
& $95.50 \pm 7.53$ & $94.54 \pm 6.35$
& $96.78 \pm 2.16$ \\

EEGNet \cite{lawhern2018eegnet}
& $89.75 \pm 5.44$ & $90.48 \pm 10.66$
& $91.32 \pm 7.94$ & $90.90 \pm 6.90$
& $94.29 \pm 3.50$
& $91.67 \pm 7.86$ & $92.12 \pm 6.77$
& $93.75 \pm 5.62$ & $92.04 \pm 8.66$
& $95.66 \pm 5.56$ \\

InceptionNet \cite{szegedy2015going}
& $86.75 \pm 7.99$ & $90.66 \pm 5.53$
& $84.96 \pm 5.30$ & $86.10 \pm 3.91$
& $94.35 \pm 1.90$
& $91.90 \pm 5.89$ & $95.00 \pm 10.12$
& $90.83 \pm 3.95$ & $91.71 \pm 5.63$
& $95.56 \pm 0.90$ \\

TSception \cite{ding2020tsception}
& $90.23 \pm 4.32$ & $89.08 \pm 4.25$
& $91.74 \pm 4.10$ & $89.90 \pm 5.65$
& $96.10 \pm 2.91$
& $91.67 \pm 5.56$ & $92.14 \pm 4.70$
& $92.08 \pm 5.84$ & $91.38 \pm 6.40$
& $96.70 \pm 4.30$ \\

GRU-Conv \cite{xu2023subject}
& $90.95 \pm 5.54$ & $90.90 \pm 4.50$
& $90.92 \pm 4.85$ & $90.68 \pm 3.23$
& $96.19 \pm 4.12$
& $92.42 \pm 6.16$ & $93.25 \pm 4.92$
& $90.83 \pm 6.87$ & $90.58 \pm 5.22$
& $97.45 \pm 5.32$ \\

DeprNet \cite{seal2021deprnet}
& $90.00 \pm 3.00$ & $90.66 \pm 5.06$
& $91.02 \pm 6.32$ & $90.84 \pm 7.39$
& $97.59 \pm 2.10$
& $91.59 \pm 6.81$ & $92.10 \pm 5.28$
& $93.75 \pm 4.62$ & $92.02 \pm 6.60$
& $98.22 \pm 1.50$ \\

GC-GRU \cite{luo2023adaptive}
& $89.74 \pm 5.64$ & $90.84 \pm 4.32$
& $87.90 \pm 3.39$ & $88.78 \pm 7.72$
& $96.01 \pm 3.62$
& $93.26 \pm 5.26$ & $93.46 \pm 6.91$
& $91.00 \pm 5.06$ & $91.79 \pm 4.59$
& $97.90 \pm 1.61$ \\

ST-GCA \cite{11369854}
& $81.80 \pm 4.94$ & $79.45 \pm 6.54$
& $84.39 \pm 5.19$ & $81.43 \pm 4.16$
& $90.40 \pm 4.75$
& $84.77 \pm 7.64$ & $81.70 \pm 6.02$
& $90.83 \pm 5.54$ & $85.40 \pm 8.75$
& $95.94 \pm 4.29$ \\

MBA-CA \cite{xu2026major}
& $86.84 \pm 4.13$ & $86.94 \pm 5.51$
& $87.04 \pm 6.76$ & $86.25 \pm 4.50$
& $92.99 \pm 6.29$
& $88.26 \pm 4.02$ & $87.52 \pm 5.13$
& $89.58 \pm 6.66$ & $87.99 \pm 4.27$
& $92.63 \pm 6.67$ \\

SC-CNN \cite{vaniya2026simplifying}
& $88.05 \pm 4.81$ & $86.89 \pm 6.98$
& $87.17 \pm 6.07$ & $86.73 \pm 5.40$
& $92.84 \pm 6.06$
& $89.92 \pm 5.23$ & $87.94 \pm 6.22$
& $89.33 \pm 7.87$ & $88.32 \pm 6.89$
& $94.90 \pm 6.40$ \\

CBraMod \cite{wang2025cbramod}
& $91.47 \pm 5.82$
& $87.94 \pm 9.26$
& \underline{$96.92 \pm 3.47$}
& $91.98 \pm 5.53$
& $93.35 \pm 6.47$
& $90.76 \pm 8.28$
& $86.40 \pm 11.09$
& \textbf{\boldmath$98.33 \pm 5.27$}
& $91.72 \pm 7.62$
& $96.87 \pm 5.14$ \\

BrainOmni-base \cite{xiao2026brainomni}
& $91.41 \pm 4.55$
& $91.03 \pm 8.70$
& $92.47 \pm 6.40$
& $91.35 \pm 4.94$
& $96.92 \pm 2.61$
& $94.17 \pm 5.62$
& $93.49 \pm 9.68$
& $96.25 \pm 8.44$
& $94.29 \pm 6.05$
& $97.61 \pm 3.83$ \\

KNN \cite{umair2025decentralized}
& $86.25 \pm 4.19$
& $82.98 \pm 8.03$
& $91.43 \pm 4.34$
& $86.80 \pm 5.11$
& $90.76 \pm 5.33$
& $92.50 \pm 7.30$
& $87.71 \pm 11.11$
& $95.83 \pm 6.50$
& $91.12 \pm 6.30$
& $94.20 \pm 4.50$ \\

\hline
\multicolumn{11}{@{}c}{\textit{Generative Augmentation Methods}} \\
\hline

DiffMDD \cite{wang2024diffmdd}
& $93.85 \pm 3.24$ & $92.52 \pm 4.98$
& $94.86 \pm 4.26$ & $93.65 \pm 3.71$
& $95.31 \pm 1.74$
& $96.55 \pm 4.30$
& \underline{$96.21 \pm 3.27$}
& $94.33 \pm 2.79$ & $95.26 \pm 4.60$
& $98.62 \pm 1.94$ \\

Hard Neg $^{*}$
& $91.50 \pm 6.05$ & $88.32 \pm 11.93$
& $95.89 \pm 6.43$ & $91.33 \pm 7.19$
& $97.58 \pm 2.11$
& $92.50 \pm 8.29$ & $89.21 \pm 14.68$
& \underline{$97.50 \pm 7.91$}
& $92.23 \pm 9.13$
& $97.40 \pm 1.79$ \\

Hybrid $^{*}$
& $92.89 \pm 9.60$ & $91.68 \pm 13.10$
& $96.70 \pm 3.61$ & $93.55 \pm 8.16$
& \underline{$98.53 \pm 1.07$}
& $95.83 \pm 10.58$ & $94.44 \pm 14.10$
& $96.67 \pm 4.85$
& $94.55 \pm 9.02$
& \underline{$98.35 \pm 1.25$} \\

Diff-Aug $^{*}$
& $93.35 \pm 4.61$ & $93.27 \pm 6.24$
& $94.19 \pm 5.67$ & $93.54 \pm 4.29$
& $97.16 \pm 2.89$
& $95.00 \pm 5.83$ & $94.92 \pm 9.46$
& $97.08 \pm 6.23$ & $95.58 \pm 5.57$
& $97.80 \pm 1.06$ \\

Clean $^{*}$
& $92.86 \pm 4.85$ & $90.91 \pm 8.31$
& $94.65 \pm 12.68$ & $91.93 \pm 7.79$
& $98.44 \pm 1.26$
& $93.26 \pm 6.60$ & $91.17 \pm 10.38$
& $95.00 \pm 15.81$ & $91.76 \pm 10.47$
& $97.85 \pm 1.80$ \\

\midrule

\textbf{F-NDD (Ours)}
& \underline{$95.20 \pm 3.63$}
& \textbf{\boldmath$94.94 \pm 4.70$}
& $95.21 \pm 10.01$
& \underline{$94.69 \pm 5.78$}
& $98.01 \pm 1.35$
& \textbf{\boldmath$97.50 \pm 4.03$}
& \textbf{\boldmath$97.22 \pm 6.00$}
& \underline{$97.50 \pm 7.91$}
& \textbf{\boldmath$97.07 \pm 5.12$}
& $98.75 \pm 1.05$ \\

\textbf{S-NDD (Ours)}
& \textbf{\boldmath$95.28 \pm 3.25$}
& \underline{$93.45 \pm 4.25$}
& \textbf{\boldmath$97.18 \pm 3.94$}
& \textbf{\boldmath$95.18 \pm 3.62$}
& \textbf{\boldmath$98.83 \pm 0.95$}
& \underline{$96.67 \pm 4.03$}
& $95.79 \pm 4.72$
& \underline{$97.50 \pm 3.66$}
& \underline{$96.30 \pm 2.97$}
& \textbf{\boldmath$98.96 \pm 0.85$} \\

\bottomrule
\multicolumn{11}{@{}l@{}}{\small $^{*}$~Augmentation strategies implemented in this study for comparison.}
\end{tabular}%
}
\end{table*}

\begin{figure*}[!t]
  \centering
\includegraphics[width=1.0\linewidth]{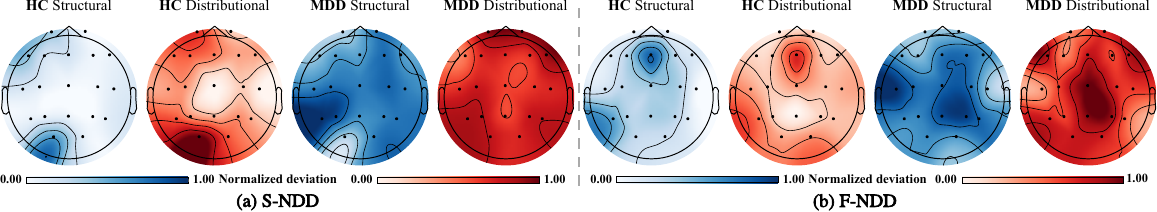}  
\vspace{-2em}
  \caption{Normative deviation visualization
for selected HC and MDD cases on Mumtaz2016
under within-dataset evaluation.
(a) S-NDD and (b) F-NDD illustrate structural
and distributional deviation patterns.
Deviation maps are normalized for visualization,
with non-comparable color scales across methods.}
\label{fig:deviation_cases1}
\vspace{-1em}
\end{figure*}

\subsection{Target HC-assisted Score Calibration.}
Cross-dataset variations in acquisition
conditions and participant populations may
shift the location and scale of diagnostic
scores. To address these shifts, we introduce
optional target HC-assisted calibration for
both S-NDD and F-NDD. Given a small set of
target HC recordings, the calibration center
and scale for each model $m$ are estimated as
\begin{align}
c_m &= Q_{\rho}\bigl(
\operatorname{logit}(p_m^{\mathrm{HC}})
\bigr),\\
s_m &= \max\bigl\{
\operatorname{IQR}(
\operatorname{logit}(p_m^{\mathrm{HC}})
),\epsilon\bigr\},
\end{align}
where $p_m^{\mathrm{HC}}$ denotes predicted
MDD probabilities of target HC recordings,
$Q_{\rho}$ is the $\rho$-quantile,
and $\epsilon$ bounds the scale.
For a target sample, the calibrated logit
is $q_m=(\operatorname{logit}(p_m)-c_m)/s_m$,
where $p_m$ is its predicted MDD probability.
The final probability is
$p_{\mathrm{final}}=
\sigma(\frac{1}{M}\sum_{m=1}^{M}q_m)$,
where $M$ is the number of models and
$\sigma$ is the sigmoid function.
Calibration uses only target HC recordings
excluded from testing, requires no target
MDD labels or model updates. Without target HC
data, predictions remain uncalibrated.

\section{Experiments}
\label{sec:experiments}

\subsection{Experimental Setup}
\label{subsec:experimental_setup}

\noindent\textbf{Implementation Details.}
EEG recordings are band-pass filtered at
0.5--50 Hz, resampled to 256 Hz, spatially
aligned to 19 electrodes, and segmented into
5-s windows with a 2.5-s stride.
The VQ-VAE codebooks contain 512 entries
of dimension 64 for S-NDD and 64 entries
of dimension 32 for F-NDD, with $\beta=0.25$.
Both methods use 500-step diffusion models
with a linear noise schedule from $10^{-4}$
to $0.02$. Normative models are trained
exclusively on source-domain HC samples
and frozen during diagnostic training.
S-NDD fuses scalar normative deviations
with deep EEG features using $\alpha=0.7$.
F-NDD extracts 456-dimensional EEG features
and estimates diffusion residuals at
$\mathcal{T}=\{25,125,250\}$ with $K=2$
noise realizations per timestep.
Both methods use a diagnostic learning rate
of $5\times10^{-4}$, with Adam for S-NDD
and F-NDD.
The auxiliary loss weight is
$\lambda_{\mathrm{dev}}=0.5$.
Further details are provided in
Appendix~\ref{app:inference}.

\noindent\textbf{Datasets.} 
We evaluate the proposed methods on Mumtaz2016~\cite{mumtaz2017electroencephalogram} and MODMA~\cite{cai2022multi}. Both datasets contain MDD and HC recordings, supporting normative modeling and cross-dataset evaluation under heterogeneous acquisition settings. Mumtaz2016 comprises 34 MDD and 30 HC subjects with 19 electrodes, while MODMA includes 24 MDD and 29 HC subjects with 128 electrodes. After preprocessing, they yield 13,953 and 6,355 EEG segments, respectively.

\noindent\textbf{Baselines.}
We compare our methods with representative
baselines from three categories.
\textit{(i) Discriminative Learning methods}
include traditional classifiers such as
KNN \shortcite{umair2025decentralized} and
XGBoost \shortcite{hassan2024impact}, as well
as deep EEG models including
EEGNet \shortcite{lawhern2018eegnet},
TSception \shortcite{ding2020tsception},
DeprNet \shortcite{seal2021deprnet},
InceptionNet \shortcite{szegedy2015going},
CWT-1D-CNN and CWT-2D-CNN
\shortcite{raghavendra2023automated},
CNN-LSTM \shortcite{ay2019automated},
CNN-Transformer \shortcite{cnn_transformer_mdd_2025},
GRU-Conv \shortcite{xu2023subject},
GC-GRU \shortcite{luo2023adaptive},
ST-GCA \shortcite{11369854},
MBA-CA \shortcite{xu2026major}, and
SC-CNN \shortcite{vaniya2026simplifying}.
We also consider
pretrained EEG models, including
CBraMod \shortcite{wang2025cbramod} and
BrainOmni-base \shortcite{xiao2026brainomni}.
\textit{(ii) Generative augmentation methods}
include DiffMDD \shortcite{wang2024diffmdd}
and four additional
comparison strategies developed in this study
(Diff-Aug,  Hard Neg, Clean, and Hybrid).
These methods investigate the use of synthetic
EEG samples for depression detection. Implementation details are provided in Appendix~\ref{sec: implementation_details}.
\textit{(iii) Domain adaptation methods}
include DANN \shortcite{ganin2016dann},
DAN \shortcite{long2015dan},
Deep CORAL \shortcite{sun2016deepcoral}, and
TFST-SDKA \shortcite{xu2026tfstsdka},
which are evaluated only in cross-dataset settings.

\begin{table}[t]
  \centering
  \caption{Cross-dataset performance (\%) on MODMA
  with Mumtaz2016 as the source.
  Results are mean $\pm$ std., with best and second-best
in bold and \underline{underlined}, respectively.}
\vspace{-1em}
\label{tab:generalization_modma}
  \begingroup
  \small
  \setlength{\tabcolsep}{2.5pt}
\renewcommand{\arraystretch}{0.95}
  \resizebox{\columnwidth}{!}{%
  \begin{tabular}{@{}lccccc@{}}
    \toprule
    \textbf{Methods}
    & \textbf{Acc ($\uparrow$)}
    & \textbf{Pre ($\uparrow$)}
    & \textbf{Rec ($\uparrow$)}
    & \textbf{F1 ($\uparrow$)}
    & \textbf{AUC ($\uparrow$)} \\

    \midrule
    \multicolumn{6}{@{}c}{\textit{Discriminative Learning Methods}} \\
    \hline

    XGBoost \shortcite{hassan2024impact}
    & $58.45 \pm 8.76$
    & $55.12 \pm 9.68$
    & $60.30 \pm 7.95$
    & $57.60 \pm 8.33$
    & $59.18 \pm 9.12$ \\

    1D-CNN-LSTM \shortcite{ay2019automated}
    & $57.20 \pm 8.42$
    & $54.35 \pm 9.15$
    & $58.12 \pm 7.63$
    & $56.18 \pm 8.84$
    & $58.14 \pm 7.92$ \\

    1D-CNN-Trans \shortcite{cnn_transformer_mdd_2025}
    & $54.51 \pm 2.69$
    & $55.90 \pm 28.45$
    & $10.35 \pm 8.23$
    & $15.77 \pm 11.17$
    & $54.23 \pm 6.87$ \\

    CWT-1D-CNN \shortcite{raghavendra2023automated}
    & $59.80 \pm 9.21$
    & $57.66 \pm 8.54$
    & $57.34 \pm 7.89$
    & $57.50 \pm 9.46$
    & $60.47 \pm 8.63$ \\

    CWT-2D-CNN \shortcite{raghavendra2023automated}
    & $56.34 \pm 5.40$
    & $56.52 \pm 27.31$
    & $11.13 \pm 12.88$
    & $16.62 \pm 16.86$
    & $52.39 \pm 5.77$ \\

    EEGNet \shortcite{lawhern2018eegnet}
    & $50.34 \pm 4.39$
    & $42.39 \pm 15.19$
    & $59.78 \pm 39.69$
    & $44.53 \pm 23.09$
    & $51.16 \pm 2.12$ \\

    InceptionNet \shortcite{szegedy2015going}
    & $53.49 \pm 3.37$
    & $45.29 \pm 15.28$
    & $8.48 \pm 5.06$
    & $13.80 \pm 7.05$
    & $49.56 \pm 3.26$ \\

    TSception \shortcite{ding2020tsception}
    & $60.12 \pm 9.53$
    & $56.45 \pm 7.78$
    & $59.85 \pm 8.92$
    & $58.10 \pm 9.16$
    & $61.05 \pm 7.48$ \\

    GRU-Conv \shortcite{xu2023subject}
    & $49.51 \pm 3.99$
    & $41.42 \pm 10.76$
    & $52.13 \pm 42.69$
    & $38.44 \pm 25.73$
    & $46.82 \pm 2.87$ \\

    DeprNet \shortcite{seal2021deprnet}
    & $56.80 \pm 8.89$
    & $52.30 \pm 9.72$
    & $58.42 \pm 8.05$
    & $55.18 \pm 7.58$
    & $56.42 \pm 8.74$ \\

    GC-GRU \shortcite{luo2023adaptive}
    & $55.40 \pm 9.34$
    & $51.20 \pm 8.67$
    & $56.75 \pm 9.08$
    & $53.82 \pm 7.91$
    & $54.89 \pm 9.26$ \\

    ST-GCA \shortcite{11369854}
    & $54.08 \pm 2.65$
    & $34.85 \pm 40.23$
    & $23.99 \pm 37.57$
    & $19.15 \pm 27.59$
    & $49.70 \pm 5.24$ \\

    MBA-CA \shortcite{xu2026major}
    & $45.94 \pm 1.77$
    & $45.42 \pm 0.54$
    & \textbf{\boldmath$95.67 \pm 10.22$}
    & $61.40 \pm 2.16$
    & $49.00 \pm 3.56$ \\

    SC-CNN \shortcite{vaniya2026simplifying}
    & $48.66 \pm 4.29$
    & $31.84 \pm 21.97$
    & $67.95 \pm 47.17$
    & $43.32 \pm 29.91$
    & $50.34 \pm 0.52$ \\

    CBraMod \cite{wang2025cbramod}
    & $46.85 \pm 2.60$
    & $45.48 \pm 0.71$
    & \underline{$85.85 \pm 21.52$}
    & $58.40 \pm 6.44$
    & $50.90 \pm 3.47$ \\

    BrainOmni-base \cite{xiao2026brainomni}
    & $54.62 \pm 1.40$
    & $49.01 \pm 2.98$
    & $48.04 \pm 22.87$
    & $45.45 \pm 17.29$
    & $55.38 \pm 2.02$ \\

    KNN \cite{umair2025decentralized}
    & $55.18 \pm 0.30$
    & $63.15 \pm 13.64$
    & $2.54 \pm 0.36$
    & $4.87 \pm 0.65$
    & $49.90 \pm 0.35$ \\

    \hline
    \multicolumn{6}{@{}c}{\textit{Generative Augmentation Methods}} \\
    \hline

    DiffMDD \shortcite{wang2024diffmdd}
    & $68.34 \pm 8.19$
    & $64.32 \pm 9.45$
    & $58.60 \pm 8.62$
    & $58.94 \pm 9.78$
    & $69.12 \pm 7.85$ \\

    Hard Neg $^{*}$
    & \underline{$69.90 \pm 10.65$}
    & $65.28 \pm 16.84$
    & $61.98 \pm 14.86$
    & $59.85 \pm 14.64$
    & \textbf{\boldmath$79.44 \pm 13.28$} \\

    Hybrid $^{*}$
    & $61.81 \pm 10.52$
    & $60.52 \pm 18.48$
    & $48.94 \pm 11.47$
    & $51.14 \pm 10.67$
    & $58.31 \pm 16.21$ \\

    Clean $^{*}$
    & $64.57 \pm 15.55$
    & $65.38 \pm 10.53$
    & $65.95 \pm 10.90$
    & $63.48 \pm 16.74$
    & $70.21 \pm 12.94$ \\

    Diff-Aug $^{*}$
    & $66.68 \pm 19.17$
    & $63.02 \pm 19.13$
    & $57.97 \pm 16.14$
    & $57.22 \pm 13.11$
    & $67.22 \pm 12.63$ \\

    \hline
    \multicolumn{6}{@{}c}{\textit{Domain Adaptation Methods}} \\
    \hline

    DANN \cite{ganin2016dann}
    & $55.81 \pm 6.98$
    & $65.01 \pm 12.31$
    & $50.00 \pm 4.17$
    & $55.95 \pm 3.27$
    & $60.01 \pm 7.66$ \\

    DAN \cite{long2015dan}
    & $51.94 \pm 1.34$
    & \underline{$75.27 \pm 21.46$}
    & $27.78 \pm 13.39$
    & $37.61 \pm 13.49$
    & $59.28 \pm 4.19$ \\

    Deep CORAL \cite{sun2016deepcoral}
    & $57.36 \pm 2.69$
    & $66.57 \pm 13.21$
    & $58.33 \pm 18.16$
    & $59.44 \pm 6.97$
    & $61.33 \pm 7.35$ \\

    TFST-SDKA \cite{xu2026tfstsdka}
    & $55.81 \pm 6.98$
    & $63.12 \pm 8.01$
    & $50.00 \pm 15.02$
    & $55.08 \pm 11.24$
    & $58.85 \pm 8.95$ \\

    \hline

      \textbf{F-NDD $^\dagger$ (Ours)}
    & $68.22 \pm 2.69$
    & \textbf{\boldmath$80.39 \pm 3.40$}
    & $56.94 \pm 2.41$
    & \underline{$66.67 \pm 2.82$}
    & $70.10 \pm 1.13$ \\

    \textbf{S-NDD (Ours)}
    & \textbf{\boldmath$78.50 \pm 5.81$}
    & $70.82 \pm 7.57$
    & $75.34 \pm 9.89$
    & \textbf{\boldmath$69.96 \pm 6.16$}
    & \underline{$79.31 \pm 3.90$} \\

    \bottomrule
    \multicolumn{6}{@{}l@{}}{\footnotesize $^{*}$~Comparison strategies developed in this study, as in Table~1. $^{\dagger}$~Target HC-assisted score calibration.}
  \end{tabular}%
  }
  \vspace{-1em}
  \endgroup
\end{table}

\begin{table}[t]
  \centering
  \caption{Cross-dataset performance (\%) on
Mumtaz2016 with MODMA as the source dataset.
All methods use target HC-assisted score
calibration. Results are mean $\pm$ std., with best and second-best
in bold and \underline{underlined}, respectively.}
  \vspace{-1em}\label{tab:reverse_transfer_hc}

  \begingroup
  \small
  \setlength{\tabcolsep}{2.5pt}
  \renewcommand{\arraystretch}{0.95}

  \resizebox{\columnwidth}{!}{%
  \begin{tabular}{@{}lccccc@{}}
    \toprule
    \textbf{Methods}
    & \textbf{Acc ($\uparrow$)}
    & \textbf{Pre ($\uparrow$)}
    & \textbf{Rec ($\uparrow$)}
    & \textbf{F1 ($\uparrow$)}
    & \textbf{AUC ($\uparrow$)} \\

    \midrule
    \multicolumn{6}{@{}c}{\textit{Discriminative Learning Methods}} \\
    \hline

    XGBoost \shortcite{hassan2024impact}
    & $48.84 \pm 8.39$
    & $76.73 \pm 7.17$
    & $30.95 \pm 17.62$
    & $41.91 \pm 20.06$
    & $57.30 \pm 3.24$ \\

    1D-CNN-LSTM \shortcite{ay2019automated}
    & $76.74 \pm 8.39$
    & $90.55 \pm 8.74$
    & $73.81 \pm 20.62$
    & $79.38 \pm 11.12$
    & $83.57 \pm 4.92$ \\

    1D-CNN-Trans \shortcite{cnn_transformer_mdd_2025}
    & $56.59 \pm 8.80$
    & $74.96 \pm 9.49$
    & $55.95 \pm 29.95$
    & $58.97 \pm 21.37$
    & $61.11 \pm 3.94$ \\

    CWT-1D-CNN \shortcite{raghavendra2023automated}
    & $75.19 \pm 1.34$
    & $82.48 \pm 5.41$
    & $79.76 \pm 11.48$
    & $80.52 \pm 2.93$
    & $80.08 \pm 5.99$ \\

    CWT-2D-CNN \shortcite{raghavendra2023automated}
    & $48.84 \pm 4.03$
    & $62.87 \pm 6.21$
    & $54.76 \pm 5.46$
    & $58.20 \pm 1.53$
    & $46.98 \pm 5.17$ \\

    EEGNet \shortcite{lawhern2018eegnet}
    & $79.07 \pm 4.03$
    & $86.30 \pm 10.25$
    & $83.33 \pm 14.43$
    & $83.60 \pm 4.11$
    & $91.11 \pm 2.16$ \\

    InceptionNet \shortcite{szegedy2015going}
    & $34.11 \pm 2.69$
    & $49.29 \pm 3.96$
    & $27.38 \pm 4.12$
    & $35.00 \pm 2.96$
    & $32.54 \pm 7.52$ \\

    TSception \shortcite{ding2020tsception}
    & $64.34 \pm 8.17$
    & $87.60 \pm 7.92$
    & $54.76 \pm 21.82$
    & $64.99 \pm 13.65$
    & $81.19 \pm 4.45$ \\

    GRU-Conv \shortcite{xu2023subject}
    & $56.59 \pm 10.49$
    & $76.12 \pm 4.13$
    & $48.81 \pm 21.82$
    & $57.38 \pm 17.93$
    & $62.94 \pm 4.16$ \\

    DeprNet \shortcite{seal2021deprnet}
    & $38.76 \pm 7.10$
    & $53.00 \pm 19.70$
    & $21.43 \pm 15.57$
    & $29.16 \pm 19.67$
    & $43.73 \pm 8.32$ \\

    GC-GRU \shortcite{luo2023adaptive}
    & $41.09 \pm 12.81$
    & $52.77 \pm 9.03$
    & $38.10 \pm 35.05$
    & $40.60 \pm 25.33$
    & $43.10 \pm 1.49$ \\

    ST-GCA \shortcite{11369854}
    & $51.94 \pm 10.49$
    & $70.75 \pm 4.17$
    & $42.86 \pm 21.43$
    & $51.67 \pm 18.46$
    & $62.70 \pm 6.28$ \\

    \shortstack[l]{MBA-CA \shortcite{xu2026major}}
    & $78.29 \pm 5.37$
    & $84.75 \pm 4.74$
    & $82.14 \pm 14.29$
    & $82.76 \pm 5.93$
    & $87.46 \pm 3.42$ \\

    \shortstack[l]{SC-CNN \shortcite{vaniya2026simplifying}}
    & $48.06 \pm 9.40$
    & $69.63 \pm 2.80$
    & $34.52 \pm 23.78$
    & $42.59 \pm 25.72$
    & $59.29 \pm 8.90$ \\

    CBraMod \cite{wang2025cbramod}
    & $43.41 \pm 3.55$
    & $69.58 \pm 7.20$
    & $23.81 \pm 8.99$
    & $34.74 \pm 10.03$
    & $52.54 \pm 4.26$ \\

    BrainOmni-base \cite{xiao2026brainomni}
    & $57.36 \pm 9.68$
    & $74.72 \pm 3.87$
    & $53.57 \pm 24.74$
    & $59.54 \pm 18.84$
    & $68.57 \pm 4.17$ \\

    KNN \cite{umair2025decentralized}
    & \underline{$83.72 \pm 2.33$}
    & \underline{$92.58 \pm 6.44$}
    & $82.14 \pm 7.14$
    & $86.73 \pm 2.23$
    & \underline{$91.67 \pm 2.27$} \\

    \hline
    \multicolumn{6}{@{}c}{\textit{Generative Augmentation methods}} \\
    \hline

    DiffMDD \shortcite{wang2024diffmdd}
    & $41.09 \pm 8.80$
    & $40.38 \pm 35.14$
    & $23.81 \pm 29.74$
    & $27.74 \pm 30.48$
    & $53.02 \pm 5.61$ \\

    Hard Neg $^{*}$
    & $37.21 \pm 2.33$
    & $57.94 \pm 8.36$
    & $14.29 \pm 0.00$
    & $22.87 \pm 0.65$
    & $38.89 \pm 4.27$ \\

    Hybrid $^{*}$
    & $40.31 \pm 3.55$
    & $72.69 \pm 23.91$
    & $16.67 \pm 2.06$
    & $26.60 \pm 1.44$
    & $65.00 \pm 11.27$ \\

    Clean $^{*}$
    & $33.33 \pm 1.34$
    & $43.33 \pm 5.77$
    & $7.14 \pm 0.00$
    & $12.25 \pm 0.22$
    & $38.17 \pm 2.76$ \\

    Diff-Aug $^{*}$
    & $37.98 \pm 1.34$
    & $65.00 \pm 8.66$
    & $10.71 \pm 0.00$
    & $18.37 \pm 0.33$
    & $37.14 \pm 3.31$ \\

    \hline
    \multicolumn{6}{@{}c}{\textit{Domain Adaptation methods}} \\
    \hline

    DANN \cite{ganin2016dann}
    & $69.77 \pm 8.39$
    & $88.47 \pm 8.39$
    & $64.29 \pm 25.00$
    & $71.78 \pm 12.08$
    & $84.21 \pm 5.11$ \\

    DAN \cite{long2015dan}
    & $51.94 \pm 23.75$
    & $45.71 \pm 40.51$
    & $42.86 \pm 49.10$
    & $42.52 \pm 42.86$
    & $72.06 \pm 22.95$ \\

    Deep CORAL \cite{sun2016deepcoral}
    & $75.97 \pm 3.55$
    & $87.10 \pm 13.90$
    & $78.57 \pm 15.57$
    & $80.87 \pm 1.48$
    & $90.40 \pm 1.62$ \\

    TFST-SDKA \cite{xu2026tfstsdka}
    & $28.68 \pm 7.10$
    & $17.78 \pm 16.78$
    & $4.76 \pm 4.12$
    & $7.43 \pm 6.46$
    & $18.89 \pm 8.12$ \\

    \hline

    \textbf{F-NDD (Ours)}
    & \textbf{\boldmath$89.15 \pm 1.34$}
    & \textbf{\boldmath$93.08 \pm 5.19$}
    & \underline{$90.48 \pm 5.46$}
    & \textbf{\boldmath$91.56 \pm 1.01$}
    & \textbf{\boldmath$94.21 \pm 0.90$} \\

    \textbf{S-NDD (Ours)}
    & $82.95 \pm 3.55$
    & $80.44 \pm 2.87$
    & \textbf{\boldmath$97.62 \pm 2.06$}
    & \underline{$88.19 \pm 2.32$}
    & $88.57 \pm 2.75$ \\

    \bottomrule
    \multicolumn{6}{@{}l@{}}{\footnotesize $^{*}$~Comparison strategies developed in this study, as in Table~1.}
  \end{tabular}%
  }
\vspace{-1em}
  \endgroup
\end{table}

\begin{figure*}[!t]
  \centering
\includegraphics[width=1.0\linewidth]{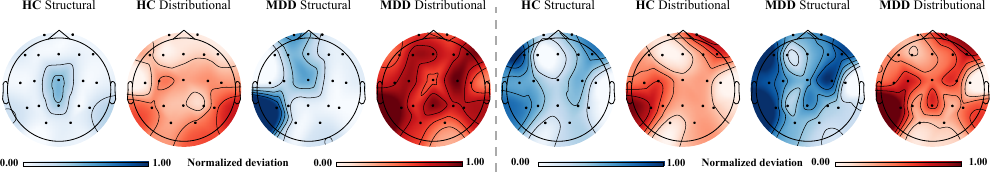}  
\vspace{-1em}
  \caption{Normative deviation visualization
for selected HC and MDD cases on MODMA 
under cross-dataset evaluation with Mumtaz2016
as the source dataset.
(a) S-NDD and (b) F-NDD illustrate structural
and distributional deviation patterns.
}
  \vspace{-1em}
  \label{fig:deviation_cases2}
\end{figure*}

\begin{figure*}[!t]
  \centering
\includegraphics[width=1.0\linewidth]{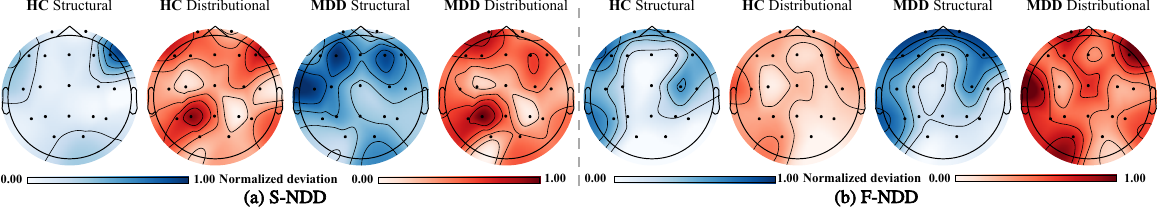}  
\vspace{-2em}
  \caption{Normative deviation visualization
for selected HC and MDD cases on Mumtaz2016
under cross-dataset evaluation with MODMA
as the source dataset.
(a) S-NDD and (b) F-NDD illustrate structural
and distributional deviation patterns.
}
  \label{fig:deviation_cases3}
  \vspace{-0.5em}
\end{figure*}


\noindent\textbf{Evaluation Metrics.}
Following established evaluation practices in
EEG-based MDD detection~\cite{wang2024diffmdd,11369854},
we report Accuracy (Acc), Precision (Pre), Recall (Rec),
F1-score (F1), and ROC-AUC (AUC) at both window and
subject levels. Both S-NDD and F-NDD generate window-level
predictions from individual EEG windows.
For S-NDD, subject-level predictions are
obtained by majority voting across windows,
with ties assigned to MDD, and subject-level
AUC is computed from the mean MDD probability.
For F-NDD, subject-level predictions are
generated directly from attention-aggregated
window representations, as described in
Section~\ref{subsec:feature}.

\subsection{MDD Detection Performance}
\label{subsec:sota_comparison}
We evaluate MDD detection on Mumtaz2016 using
subject-independent 10-fold cross-validation,
with all windows from each subject assigned to the same fold.
As shown in Table~\ref{tab:sota_comparison},
S-NDD achieves the highest window-level accuracy,
recall, F1, and AUC of 95.28\%, 97.18\%, 95.18\%,
and 98.83\%, respectively.
Its accuracy exceeds CWT-1D-CNN, the strongest
discriminative baseline in accuracy, by 3.74 percentage points.
This improvement is accompanied by high sensitivity
without a substantial loss of precision.
For example, CBraMod achieves a comparable recall
of 96.92\% but only 87.94\% precision, whereas S-NDD
attains 97.18\% recall and 93.45\% precision,
indicating better control of false positives.
S-NDD and F-NDD exhibit different strengths across
evaluation levels.
Their window-level accuracies are close
(95.28\% and 95.20\%, respectively), while S-NDD
achieves higher recall, F1, and AUC.
At the subject level, F-NDD achieves the highest
accuracy, precision, and F1 of 97.50\%, 97.22\%,
and 97.07\%, respectively.
Both variants attain 97.50\% recall, with F-NDD
achieving higher precision and S-NDD obtaining
the highest AUC of 98.96\%.
These results support the diagnostic value of normative
deviations across both evaluation levels, with S-NDD
providing stronger window-level discrimination and
F-NDD achieving better subject-level accuracy and F1.

\noindent\textbf{Beyond Generative Augmentation.}
Unlike generative augmentation methods that expand training
data with synthetic EEG samples, S-NDD and F-NDD use
generative models to establish healthy references and
estimate normative deviations.
As shown in Table~\ref{tab:sota_comparison}, both variants
outperform DiffMDD and the four augmentation baselines
in accuracy and F1 at both evaluation levels.
At the window level, S-NDD achieves 95.28\% accuracy
and 95.18\% F1, improving over DiffMDD by 1.43 and
1.53 percentage points, respectively.
Although Hybrid achieves a competitive AUC of 98.53\%,
its accuracy remains at 92.89\%, suggesting that strong
ranking performance does not necessarily translate
into equally effective classification.
S-NDD achieves a higher AUC of 98.83\% together with
higher accuracy and F1.
At the subject level, F-NDD achieves the highest accuracy
and F1 of 97.50\% and 97.07\%, while S-NDD obtains
the highest AUC of 98.96\%.
These results support learning deviations from healthy
references as an effective alternative to synthetic
data augmentation for EEG-based MDD detection.

\noindent\textbf{Normative Deviation Visualization.}
Figure~\ref{fig:deviation_cases1} visualizes
structural and distributional deviations
for selected HC and MDD cases on Mumtaz2016.
In both S-NDD and F-NDD, the selected MDD
cases exhibit more widespread structural
deviations than their HC counterparts,
suggesting that healthy-reference
reconstruction captures differences
between the diagnostic groups.
Meanwhile, distributional deviations
exhibit distinct spatial patterns,
indicating that they characterize
different aspects of EEG variations
from structural reconstruction errors.
The two formulations also produce
different deviation maps, reflecting
their signal- and feature-space
representations, respectively.
Together, these observations illustrate
how healthy references characterize
EEG variations at different levels,
supporting the use of normative deviations
in diagnostic learning.

\begin{figure}[t]
  \centering
  \includegraphics[width=\columnwidth]
  {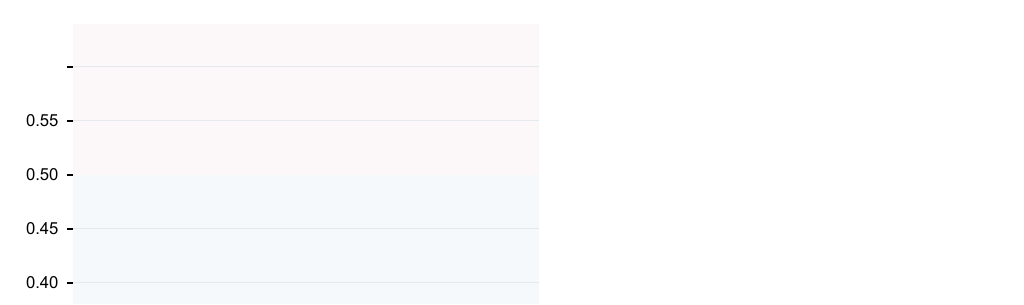}
    \vspace{-1em}
  \caption{F-NDD scores before and after
target HC calibration for (a) HC and (b) MDD.
Green lines indicate corrected predictions,
crosses denote errors, and the dashed line
marks the decision threshold.}
\label{fig:fndd_score_raincloud}
  \vspace{-1em}
\end{figure}

\subsection{Cross-Dataset Generalization}
\label{subsec:generalization}
\noindent\textbf{Mumtaz2016 to MODMA.}
We train S-NDD and F-NDD on Mumtaz2016 and evaluate them
on MODMA, mapping its 128-electrode recordings to
19 electrodes through spatial adaptation.
F-NDD additionally uses target-domain HC data for score calibration.
As shown in Table~\ref{tab:generalization_modma},
S-NDD achieves the highest accuracy of 78.50\% and
F1 of 69.96\%.
Compared with the best results among DANN, DAN,
Deep CORAL, and TFST-SDKA, these represent gains
of 21.14 and 10.52 percentage points, respectively.
S-NDD also exceeds the strongest non-ours results
in accuracy and F1 by 8.60 and 6.48 points,
while achieving an AUC comparable to the best baseline
(79.31\% versus 79.44\%).
These results support the effectiveness of normative
deviation modeling beyond the evaluated domain
alignment and augmentation approaches.
Compared with calibrated F-NDD, S-NDD achieves higher
recall (75.34\% versus 56.94\%) with lower precision,
yielding better overall accuracy and F1 without
target-domain calibration.

\noindent\textbf{MODMA to Mumtaz2016.}
To account for potential shifts in prediction scores,
we evaluate reverse transfer using target HC-assisted
score calibration for all methods.
Seven target HC record groups are reserved for calibration
and excluded from diagnostic testing, with model parameters
kept fixed.
Table~\ref{tab:reverse_transfer_hc} shows that F-NDD
achieves the highest accuracy, precision, F1, and AUC
of 89.15\%, 93.08\%, 91.56\%, and 94.21\%, respectively.
It outperforms KNN, the strongest baseline in accuracy,
by 5.43 percentage points in accuracy and 4.83 points in F1.
Its accuracy also exceeds Deep CORAL, the strongest
domain adaptation baseline, by 13.18 points.
S-NDD achieves the highest recall of 97.62\%,
whereas F-NDD combines 90.48\% recall with substantially
higher precision (93.08\% versus 80.44\%).
Thus, under the same calibration setting, F-NDD provides
a more favorable balance between MDD detection and
false-positive control.

\noindent\textbf{Visualization Analysis.}
Figure~\ref{fig:deviation_cases2}
and Figure~\ref{fig:deviation_cases3} visualize structural
and distributional deviations in both transfer directions.
The selected cases exhibit different regional patterns
across the two views, illustrating how they characterize
distinct aspects of departure from the healthy reference.
For example, the S-NDD MDD case on MODMA exhibits
localized structural deviations alongside more widespread
distributional deviations.
Figure~\ref{fig:fndd_score_raincloud} further illustrates
the effect of HC-assisted calibration on F-NDD.
Most HC scores move below the threshold of 0.5,
correcting false positives, while most MDD scores remain
above it, with a few additional false negatives.
Correspondingly, accuracy increases from 67.44\% to
89.15\% and F1 from 79.42\% to 91.56\%, whereas AUC
changes modestly from 93.41\% to 94.21\%.
This suggests that calibration primarily improves
threshold-based decisions while largely preserving
the model's discriminative ranking.
\begin{table}[t]
  \centering
  
\caption{Ablation results  (\%) of S-NDD and F-NDD
on Mumtaz2016. Results are reported
as mean $\pm$ std., with the best and
second-best results in bold and underlined,
respectively.}
\vspace{-1em}
  \label{tab:ablation_combined}

  \begingroup
  \small
  \setlength{\tabcolsep}{2.5pt}
  \renewcommand{\arraystretch}{0.95}

  \resizebox{\columnwidth}{!}{%
  \begin{tabular}{@{}lccccc@{}}
    \toprule
    \textbf{Methods}
    & \textbf{Acc ($\uparrow$)}
    & \textbf{Pre ($\uparrow$)}
    & \textbf{Rec ($\uparrow$)}
    & \textbf{F1 ($\uparrow$)}
    & \textbf{AUC ($\uparrow$)} \\

    \midrule
    \multicolumn{6}{c}{\textit{S-NDD Ablation Study}} \\
    \midrule

    w/o Both
    & $85.56 \pm 4.61$
    & $84.90 \pm 6.35$
    & $87.97 \pm 5.76$
    & $84.98 \pm 4.36$
    & $92.39 \pm 5.03$ \\

    w/o Dist.
    & $90.40 \pm 5.94$
    & $90.69 \pm 5.99$
    & $92.72 \pm 5.72$
    & $90.91 \pm 5.05$
    & $98.24 \pm 1.67$ \\

    w/o Struct.
    & $93.11 \pm 4.85$
    & $92.38 \pm 7.72$
    & $94.34 \pm 8.09$
    & $92.95 \pm 4.84$
    & $97.92 \pm 2.96$ \\

    \addlinespace[2pt]

    Dec. Fusion
    & \underline{$93.12 \pm 5.05$}
    & \textbf{\boldmath$94.10 \pm 7.49$}
    & $91.62 \pm 7.60$
    & $92.00 \pm 8.41$
    & \underline{$98.53 \pm 1.44$} \\

    \addlinespace[2pt]

    Mahalanobis
    & $92.62 \pm 5.01$
    & $90.91 \pm 7.59$
    & $95.31 \pm 5.25$
    & \underline{$93.06 \pm 6.72$}
    & $97.84 \pm 2.97$ \\

    OCSVM
    & $92.25 \pm 6.23$
    & $88.66 \pm 7.28$
    & \textbf{\boldmath$97.43 \pm 2.72$}
    & $92.84 \pm 5.76$
    & $97.94 \pm 3.58$ \\

    Simple VAE
    & $92.56 \pm 4.80$
    & $91.97 \pm 6.89$
    & $92.71 \pm 8.55$
    & $92.34 \pm 5.78$
    & $97.61 \pm 3.62$ \\

    \midrule

    \textbf{S-NDD (Ours)}
    & \textbf{\boldmath$95.28 \pm 3.25$}
    & \underline{$93.45 \pm 4.25$}
    & \underline{$97.18 \pm 3.94$}
    & \textbf{\boldmath$95.18 \pm 3.62$}
    & \textbf{\boldmath$98.83 \pm 0.95$} \\

    \midrule
    \multicolumn{6}{c}{\textit{F-NDD Ablation Study}} \\
    \midrule

    w/o Deviations
    & \underline{$88.75 \pm 4.96$}
    & \underline{$89.43 \pm 7.58$}
    & $89.59 \pm 6.41$
    & \underline{$89.15 \pm 4.07$}
    & \underline{$95.15 \pm 3.90$} \\

    w/o Struct.
    & $87.12 \pm 5.13$
    & $86.62 \pm 10.45$
    & $89.64 \pm 6.13$
    & $87.56 \pm 5.27$
    & $94.08 \pm 3.49$ \\

    w/o Dist.
    & $87.42 \pm 6.94$
    & $85.34 \pm 12.61$
    & $90.01 \pm 5.38$
    & $87.28 \pm 8.36$
    & $92.88 \pm 7.09$ \\

    \addlinespace[2pt]

    Spectral only (95-D)
    & $86.11 \pm 6.74$
    & $85.15 \pm 10.26$
    & $86.70 \pm 10.42$
    & $85.56 \pm 9.00$
    & $92.93 \pm 5.99$ \\

    Correlation only (361-D)
    & $74.15 \pm 6.10$
    & $74.42 \pm 13.15$
    & $78.14 \pm 9.25$
    & $75.01 \pm 6.42$
    & $82.37 \pm 7.37$ \\

    \midrule

    \textbf{F-NDD (Ours)}
    & \textbf{\boldmath$95.20 \pm 3.63$}
    & \textbf{\boldmath$94.94 \pm 4.70$}
    & \textbf{\boldmath$95.21 \pm 10.01$}
    & \textbf{\boldmath$94.69 \pm 5.78$}
    & \textbf{\boldmath$98.01 \pm 1.35$} \\

    \bottomrule
  \end{tabular}%
  }
\vspace{-1em}
  \endgroup
\end{table}

\subsection{Ablation Study}
\label{subsec:ablation}
\noindent\textbf{Scalar-level Deviation (S-NDD).}
As shown in Table~\ref{tab:ablation_combined},
removing both structural and distributional
deviations reduces accuracy from 95.28\%
to 85.56\%, indicating the contribution
of normative information beyond deep EEG
representations. Removing the distributional
or structural component yields 90.40\%
and 93.11\% accuracy, respectively,
with a larger reduction observed when
distributional deviations are excluded.
Replacing deviation-guided fusion with
decision-level fusion decreases accuracy
to 93.12\%, supporting the adopted
feature-fusion strategy.
Alternative normative priors based on
Mahalanobis distance, OCSVM, and Simple VAE
achieve accuracies of 92.62\%, 92.25\%,
and 92.56\%, respectively.
Although OCSVM attains slightly higher
recall (97.43\% versus 97.18\%),
S-NDD achieves the highest accuracy,
F1, and AUC among these configurations.
These results support the combined use
of structural and distributional references
with deviation-guided feature fusion.

\noindent\textbf{Feature-level Deviation (F-NDD).}
As shown in Table~\ref{tab:ablation_combined},
removing both deviation components reduces
accuracy from 95.20\% to 88.75\%.
Excluding structural or distributional
residuals further reduces accuracy to
87.12\% and 87.42\%, respectively,
indicating that the two components
contribute jointly to diagnostic performance.
The larger degradation relative to
removing both deviations also suggests
that retaining only one residual component
does not necessarily improve the learned
representation. Regarding feature composition,
using only spectral features (95-D) yields
86.11\% accuracy, while correlation-only
features (361-D) achieve 74.15\%.
The complete F-NDD achieves 95.20\%
accuracy and 94.69\% F1, supporting
the combined use of spectral and
inter-electrode correlation information
for feature-level deviation modeling.
Overall, the ablations highlight the
contribution of structural and distributional
deviations in both diagnostic formulations.

\begin{figure}[t]
    \centering
    \includegraphics[width=\linewidth]{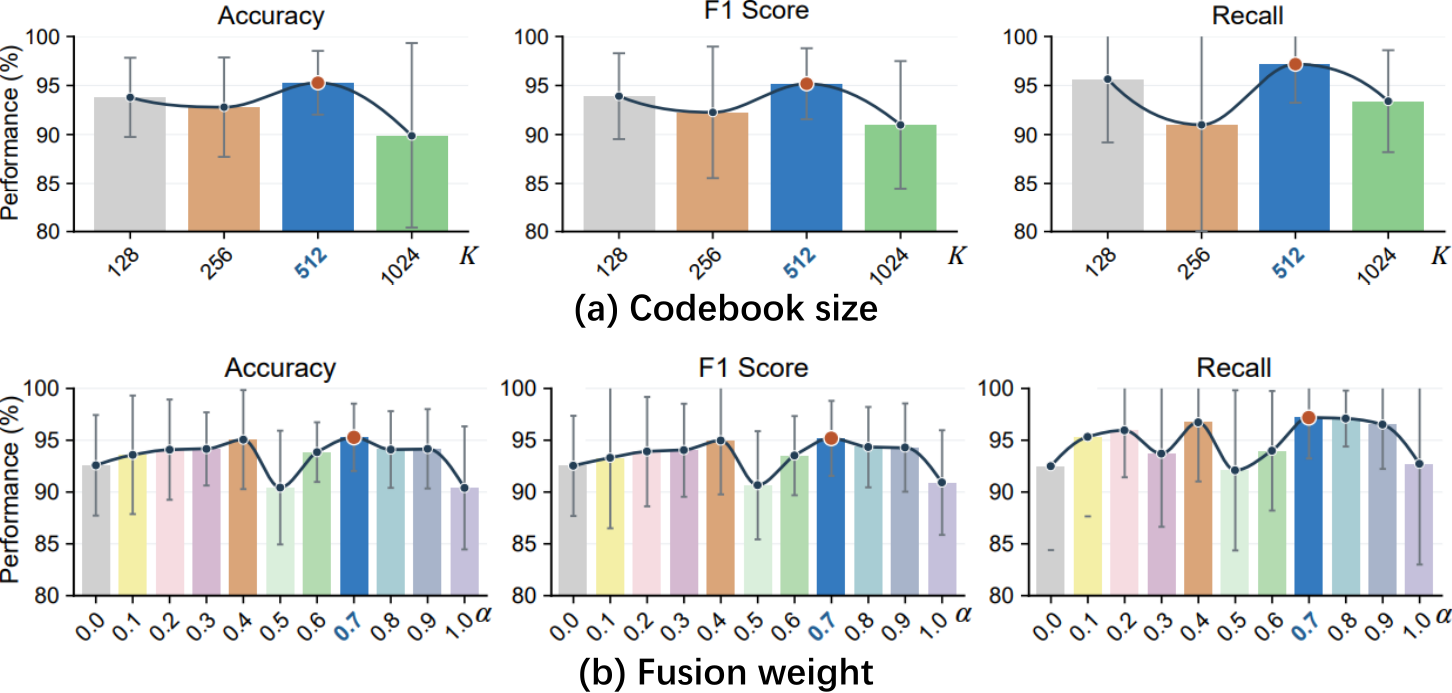}
    \vspace{-1.8em}
    \caption{Sensitivity of S-NDD to
(a) codebook size $K$ and (b) fusion weight
$\alpha$ on Mumtaz2016.  Orange dots
indicating the highest means.}
   \vspace{-1em}
    \label{fig:hyperparameter}
\end{figure}
\begin{figure}[t]
    \centering
    \includegraphics[width=\linewidth]{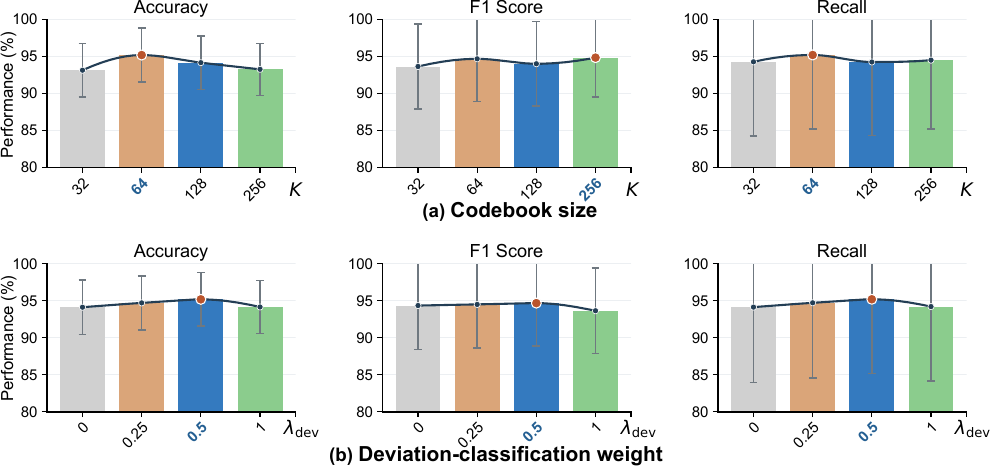}
     \vspace{-2em}
    \caption{Sensitivity of F-NDD to
(a) codebook size $K$ and (b) auxiliary
loss weight $\lambda_{\mathrm{dev}}$
on Mumtaz2016. Orange dots
indicating the highest means.}
     \vspace{-1em}\label{fig:fndd_sensitivity}
\end{figure}

\subsection{Sensitivity Analysis}
\label{subsec:hyperparameter}

We examine the sensitivity of S-NDD and
F-NDD to their key hyperparameters on
Mumtaz2016. As shown in
Fig.~\ref{fig:hyperparameter}, S-NDD
achieves the highest mean accuracy,
F1, and recall at $K=512$ and
$\alpha=0.7$ in the respective sweeps.
Performance generally remains stable
across moderate fusion weights, while
larger codebooks do not consistently
improve the results.
Figure~\ref{fig:fndd_sensitivity}
shows that F-NDD achieves the highest
mean accuracy and recall at $K=64$,
while $K=256$ yields the highest F1.
For the auxiliary loss weight,
$\lambda_{\mathrm{dev}}=0.5$ achieves
the highest means across all three
metrics. Overall, F-NDD exhibits small performance variations
across the tested configurations.

\section{Conclusion and Future Work}
In this paper, we investigate healthy
normative deviations for EEG-based MDD
detection through two approaches, S-NDD
and F-NDD. Both learn structural and
distributional references from healthy
controls and incorporate deviations
at scalar and feature levels, respectively.
Experiments on two public EEG datasets
demonstrate their effectiveness under
within-source and bidirectional cross-source
settings. The two approaches exhibit
distinct strengths across diagnostic levels
and transfer directions. Optional target HC-assisted
score calibration further improves
performance in selected cross-source
settings. These findings highlight the
value of healthy normative references
in complementing supervised EEG representations.
As spatial interpolation mainly aligns
electrode geometry, future work will
investigate unsupervised domain adaptation
to mitigate remaining acquisition-related
differences across datasets.

\bibliographystyle{ACM-Reference-Format}
\bibliography{ref/sample-base}

@article{globaldisability,
  title={Global and regional quality of care index in major depressive disorder: the global burden of disease study 2021},
  author={Xu, Chenyue and Zhu, Longzhu and Lai, Jianbo and Luo, Zeyu and Ying, Jiayao and Hu, Shaohua and Song, Peige and Yang, Jianzhong},
  journal={International Journal for Equity in Health},
  year={2026},
  publisher={Springer}
}

@article{EEG,
  title={Machine learning-assisted resting-state electroencephalography improves diagnostic accuracy in psychiatric disorders: A narrative review},
  author={Leccisotti, Ivana and Mollica, Anita and Laurello, Rossana and Moretti, Maria Claudia and Altamura, Mario and Bellomo, Antonello and Panza, Francesco and Lozupone, Madia},
  journal={Advanced Technology in Neuroscience},
  volume={3},
  number={1},
  pages={21--33},
  year={2026},
  publisher={Medknow}
}

@article{xu2026major,
  title={A major depressive disorder detection network incorporating multi-branch attention and center alignment},
  author={Xu, Chen-Yang and Li, Wen-Qiang and Zhang, Lan and Zhang, Yong-Hui and Meng, Qing-Hao},
  journal={Biomedical Signal Processing and Control},
  volume={125},
  pages={110728},
  year={2026},
  publisher={Elsevier}
}

@article{khayretdinova2024prediction,
  title={Prediction of brain sex from EEG: using large-scale heterogeneous dataset for developing a highly accurate and interpretable ML model},
  author={Khayretdinova, Mariam and Zakharov, Ilya and Pshonkovskaya, Polina and Adamovich, Timothy and Kiryasov, Andrey and Zhdanov, Andrey and Shovkun, Alexey},
  journal={NeuroImage},
  volume={285},
  pages={120495},
  year={2024},
  publisher={Elsevier}
}

@article{choudhary2024spatio,
  title={Spatio-temporal features based deep learning model for depression detection using two electrodes},
  author={Choudhary, Shubham and Kumar Bajpai, Manish and Kumari Bharti, Kusum},
  journal={Measurement Science and Technology},
  volume={35},
  number={8},
  pages={086015},
  year={2024},
  publisher={IOP Publishing}
}

@article{EEGSignal,
  title={Deep learning approaches for EEG-based healthcare applications: a comprehensive review},
  author={Lyu, RuiFang},
  journal={Frontiers in Human Neuroscience},
  volume={19},
  pages={1689073},
  year={2026}
}

@incollection{lawal2026transfer,
  title={Transfer learning and domain adaptation in neuroinformatics},
  author={Lawal, Mustapha Abdulrahman and Abayomi, Abdultaofeek and Salihu, Abubakar Abba and Aliyu, Anes Aihong and Aliyu, Muazu Aminu and Yakubu, Ismail Zahraddeen},
  booktitle={Deep Learning Applications in Neuroinformatics},
  pages={287--310},
  year={2026},
  publisher={Elsevier}
}

@article{choi2026fail,
  title={Fail closed trust gated synthetic augmentation governs tail risk under subject shift in EEG},
  author={Choi, Daniel and Yip, Cordelia and Choi, Andrew and Park, Junho},
  journal={bioRxiv},
  pages={2026--01},
  year={2026},
  publisher={Cold Spring Harbor Laboratory}
}

@article{van2017neural,
  title={Neural discrete representation learning},
  author={Van Den Oord, Aaron and Vinyals, Oriol and others},
  journal={Advances in neural information processing systems},
  volume={30},
  year={2017}
}

@article{yang2025thd,
  title={THD-BAR: Topology Hierarchical Derived Brain Autoregressive Modeling for EEG Generic Representations},
  author={Yang, Wenchao and Yan, Weidong and Liu, Wenkang and Ma, Yulan and Li, Yang},
  journal={arXiv preprint arXiv:2511.13733},
  year={2025}
}

@article{ho2020denoising,
  title={Denoising diffusion probabilistic models},
  author={Ho, Jonathan and Jain, Ajay and Abbeel, Pieter},
  journal={Advances in neural information processing systems},
  volume={33},
  pages={6840--6851},
  year={2020}
}

@inproceedings{pinaya2022fast,
  title={Fast unsupervised brain anomaly detection and segmentation with diffusion models},
  author={Pinaya, Walter HL and Graham, Mark S and Gray, Robert and Da Costa, Pedro F and Tudosiu, Petru-Daniel and Wright, Paul and Mah, Yee H and MacKinnon, Andrew D and Teo, James T and Jager, Rolf and others},
  booktitle={International Conference on Medical Image Computing and Computer-Assisted Intervention},
  pages={705--714},
  year={2022},
  organization={Springer}
}

@article{rutherford2022normative,
  title={The normative modeling framework for computational psychiatry},
  author={Rutherford, Saige and Kia, Seyed Mostafa and Wolfers, Thomas and Fraza, Charlotte and Zabihi, Mariam and Dinga, Richard and Berthet, Pierre and Worker, Amanda and Verdi, Serena and Ruhe, Henricus G and others},
  journal={Nature protocols},
  volume={17},
  number={7},
  pages={1711--1734},
  year={2022},
  publisher={Nature Publishing Group UK London}
}

@article{mastgcn_mdd,
  title={Mast-gcn: Multi-scale adaptive spatial-temporal graph convolutional network for eeg-based depression recognition},
  author={Lu, Haifeng and You, Zhiyang and Guo, Yi and Hu, Xiping},
  journal={IEEE Transactions on Affective Computing},
  volume={15},
  number={4},
  pages={1985--1996},
  year={2024},
  publisher={IEEE}
}

@inproceedings{cnn_transformer_mdd_2025,
  title={Enhancing Depression Detection from Emotion EEG with Temporal-Spatial-Spectral Representation Learning},
  author={Peng, Dan and Zheng, Wei-Long and Lu, Bao-Liang},
  booktitle={2025 47th Annual International Conference of the IEEE Engineering in Medicine and Biology Society (EMBC)},
  pages={1--5},
  year={2025},
  organization={IEEE}
}

@article{singh2024slitranet,
  title={Slitranet: an EEG-based automated diagnosis framework for major depressive disorder monitoring using a novel LGCN and transformer-based hybrid deep learning approach},
  author={Singh, Anurag and Tiwari, Vivek and Patel, Harshita and Vivekananda, GN and Rajput, Dharmendra Singh and others},
  journal={IEEE Access},
  volume={12},
  pages={173109--173126},
  year={2024},
  publisher={IEEE}
}

@article{hou2025lightweight,
  title={A lightweight convolutional transformer neural network for EEG-based depression recognition},
  author={Hou, Pengfei and Li, Xiaowei and Zhu, Jing and Hu, Bin},
  journal={Biomedical Signal Processing and Control},
  volume={100},
  pages={107112},
  year={2025},
  publisher={Elsevier}
}

@ARTICLE{11369854,
  author={Sun, Yubing and Sun, Jiaqi and Zhou, Zijian and Cai, Jing and Cui, Wenjie and Liu, Guangda},
  journal={IEEE Sensors Journal}, 
  title={A Novel Spatial-Temporal Graph Neural Network for Major Depressive Disorder Detection Based on Resting State EEG Signals}, 
  year={2026},
  volume={26},
  number={6},
  pages={8660-8671},
  doi={10.1109/JSEN.2026.3657750}}

@article{vaniya2026simplifying,
  title={Simplifying Depression Diagnosis: Single-Channel EEG and Deep Learning Approaches},
  author={Vaniya, Shruthi Narayanan and Habib, Ahsan and Angelova, Maia and Karmakar, Chandan},
  journal={IEEE Journal of Biomedical and Health Informatics},
  year={2026},
  publisher={IEEE}
}

@article{olbrich2026deep,
  title={Deep learning using electroencephalogram (EEG) data for diagnosing and predicting SSRI response in major depressive disorder},
  author={Olbrich, Sebastian and Jaworska, Natalia and de la Salle, Sara and Knott, Verner and Blier, Pierre and Brunovsky, Martin and Welt, Tobias and de Bardeci, Mateo and Teng-Ip, Cheng},
  journal={Communications Medicine},
  volume={6},
  number={1},
  pages={159},
  year={2026},
  publisher={Nature Publishing Group UK London}
}

@inproceedings{zhao2024eeg,
  title={EEG data augmentation for emotion recognition using diffusion model},
  author={Zhao, Yi-Dong and Liu, Yan-Kai and Zheng, Wei-Long and Lu, Bao-Liang},
  booktitle={2024 46th annual international conference of the ieee engineering in medicine and biology society (embc)},
  pages={1--4},
  year={2024},
  organization={IEEE}
}

@article{klem1999ten,
  title={The ten-twenty electrode system of the international federation. The international federation of clinical neurophysiology},
  author={Klem, George H},
  journal={Electroencephalogr. Clin. Neurophysiol. Suppl.},
  volume={52},
  pages={3--6},
  year={1999}
}

@article{mumtaz2017electroencephalogram,
  title={Electroencephalogram (EEG)-based computer-aided technique to diagnose major depressive disorder (MDD)},
  author={Mumtaz, Wajid and Xia, Likun and Ali, Syed Saad Azhar and Yasin, Mohd Azhar Mohd and Hussain, Muhammad and Malik, Aamir Saeed},
  journal={Biomedical Signal Processing and Control},
  volume={31},
  pages={108--115},
  year={2017},
  publisher={Elsevier}
}

@article{cai2022multi,
  title={A multi-modal open dataset for mental-disorder analysis},
  author={Cai, Hanshu and Yuan, Zhenqin and Gao, Yiwen and Sun, Shuting and Li, Na and Tian, Fuze and Xiao, Han and Li, Jianxiu and Yang, Zhengwu and Li, Xiaowei and others},
  journal={Scientific data},
  volume={9},
  number={1},
  pages={178},
  year={2022},
  publisher={Nature Publishing Group UK London}
}

@article{ay2019automated,
  title={Automated depression detection using deep representation and sequence learning with EEG signals},
  author={Ay, Betul and Yildirim, Ozal and Talo, Muhammed and Baloglu, Ulas Baran and Aydin, Galip and Puthankattil, Subha D and Acharya, U Rajendra},
  journal={Journal of medical systems},
  volume={43},
  number={7},
  pages={205},
  year={2019},
  publisher={Springer}
}

@article{raghavendra2023automated,
  title={Automated detection and screening of depression using continuous wavelet transform with electroencephalogram signals},
  author={Raghavendra, U and Gudigar, Anjan and Chakole, Yashas and Kasula, Praneet and Subha, DP and Kadri, Nahrizul Adib and Ciaccio, Edward J and Acharya, U Rajendra},
  journal={Expert Systems},
  volume={40},
  number={4},
  pages={e12803},
  year={2023},
  publisher={Wiley Online Library}
}

@article{lawhern2018eegnet,
  title={EEGNet: a compact convolutional neural network for EEG-based brain--computer interfaces},
  author={Lawhern, Vernon J and Solon, Amelia J and Waytowich, Nicholas R and Gordon, Stephen M and Hung, Chou P and Lance, Brent J},
  journal={Journal of neural engineering},
  volume={15},
  number={5},
  pages={056013},
  year={2018},
  publisher={iOP Publishing}
}

@inproceedings{szegedy2015going,
  title={Going deeper with convolutions},
  author={Szegedy, Christian and Liu, Wei and Jia, Yangqing and Sermanet, Pierre and Reed, Scott and Anguelov, Dragomir and Erhan, Dumitru and Vanhoucke, Vincent and Rabinovich, Andrew},
  booktitle={Proceedings of the IEEE conference on computer vision and pattern recognition},
  pages={1--9},
  year={2015}
}

@inproceedings{ding2020tsception,
  title={TSception: A deep learning framework for emotion detection using EEG},
  author={Ding, Yi and Robinson, Neethu and Zeng, Qiuhao and Chen, Duo and Wai, Aung Aung Phyo and Lee, Tih-Shih and Guan, Cuntai},
  booktitle={2020 international joint conference on neural networks (IJCNN)},
  pages={1--7},
  year={2020},
  organization={IEEE}
}

@article{wang2024diffmdd,
  title={Diffmdd: A diffusion-based deep learning framework for mdd diagnosis using eeg},
  author={Wang, Yilin and Zhao, Sha and Jiang, Haiteng and Li, Shijian and Luo, Benyan and Li, Tao and Pan, Gang},
  journal={IEEE Transactions on Neural Systems and Rehabilitation Engineering},
  volume={32},
  pages={728--738},
  year={2024},
  publisher={IEEE}
}

@article{xu2023subject,
  title={Subject-independent EEG emotion recognition with hybrid spatio-temporal GRU-Conv architecture},
  author={Xu, Guixun and Guo, Wenhui and Wang, Yanjiang},
  journal={Medical \& Biological Engineering \& Computing},
  volume={61},
  number={1},
  pages={61--73},
  year={2023},
  publisher={Springer}
}

@article{seal2021deprnet,
  title={DeprNet: A deep convolution neural network framework for detecting depression using EEG},
  author={Seal, Ayan and Bajpai, Rishabh and Agnihotri, Jagriti and Yazidi, Anis and Herrera-Viedma, Enrique and Krejcar, Ondrej},
  journal={IEEE Transactions on Instrumentation and Measurement},
  volume={70},
  pages={1--13},
  year={2021},
  publisher={IEEE}
}

@article{hassan2024impact,
  title={Impact of feature selection techniques on the performance of machine learning models for depression detection using EEG data},
  author={Hassan, Marwa and Kaabouch, Naima},
  journal={Applied Sciences},
  volume={14},
  number={22},
  pages={10532},
  year={2024},
  publisher={MDPI}
}

@article{leuchter2012resting,
  title={Resting-state quantitative electroencephalography reveals increased neurophysiologic connectivity in depression},
  author={Leuchter, Andrew F and Cook, Ian A and Hunter, Aimee M and Cai, Chaochao and Horvath, Steve},
  journal={PloS one},
  volume={7},
  number={2},
  pages={e32508},
  year={2012},
  publisher={Public Library of Science San Francisco, USA}
}

@article{das2020resting,
  title={Resting state quantitative electroencephalogram power spectra in patients with depressive disorder as compared to normal controls: an observational study},
  author={Das, Jnanamay and Yadav, Shailly},
  journal={Indian Journal of Psychological Medicine},
  volume={42},
  number={1},
  pages={30--38},
  year={2020},
  publisher={SAGE Publications Sage India: New Delhi, India}
}

@article{marquand2019conceptualizing,
  title={Conceptualizing mental disorders as deviations from normative functioning},
  author={Marquand, Andre F and Kia, Seyed Mostafa and Zabihi, Mariam and Wolfers, Thomas and Buitelaar, Jan K and Beckmann, Christian F},
  journal={Molecular Psychiatry},
  volume={24},
  number={10},
  pages={1415--1424},
  year={2019}
}

@article{acharya2018depression,
  author = {Acharya, U. Rajendra and Oh, Shu Lih and Hagiwara, Yuki and Tan, Jen Hong and Adeli, Hojjat and Subha, D. P.},
  title = {Automated {EEG}-based screening of depression using deep convolutional neural network},
  journal = {Computer Methods and Programs in Biomedicine},
  year = {2018}, volume = {161}, pages = {103--113},
  doi = {10.1016/j.cmpb.2018.04.012}
}

@article{luo2023adaptive,
  author = {Luo, Gang and Rao, Hong and An, Panfeng and Li, Yunxia and Hong, Ruiyun and Chen, Wenwu and Chen, Shengbo},
  title = {Exploring Adaptive Graph Topologies and Temporal Graph Networks for {EEG}-Based Depression Detection},
  journal = {IEEE Transactions on Neural Systems and Rehabilitation Engineering},
  year = {2023}, volume = {31}, pages = {3947--3957},
  doi = {10.1109/TNSRE.2023.3320693}
}

@article{song2023eegconformer,
  author = {Song, Yonghao and Zheng, Qingqing and Liu, Bingchuan and Gao, Xiaorong},
  title = {{EEG} Conformer: Convolutional Transformer for {EEG} Decoding and Visualization},
  journal = {IEEE Transactions on Neural Systems and Rehabilitation Engineering},
  year = {2023}, volume = {31}, pages = {710--719},
  doi = {10.1109/TNSRE.2022.3230250}
}

@article{ganin2016dann,
  author = {Ganin, Yaroslav and Ustinova, Evgeniya and Ajakan, Hana and Germain, Pascal and Larochelle, Hugo and Laviolette, Francois and March, Mario and Lempitsky, Victor},
  title = {Domain-Adversarial Training of Neural Networks},
  journal = {Journal of Machine Learning Research},
  year = {2016}, volume = {17}, number = {59}, pages = {1--35},
  url = {https://jmlr.org/papers/v17/15-239.html}
}

@inproceedings{long2015dan,
  author = {Long, Mingsheng and Cao, Yue and Wang, Jianmin and Jordan, Michael I.},
  title = {Learning Transferable Features with Deep Adaptation Networks},
  booktitle = {Proceedings of the 32nd International Conference on Machine Learning},
  series = {Proceedings of Machine Learning Research},
  year = {2015}, volume = {37}, pages = {97--105},
  url = {https://proceedings.mlr.press/v37/long15.html}
}

@inproceedings{sun2016deepcoral,
  author = {Sun, Baochen and Saenko, Kate},
  title = {Deep {CORAL}: Correlation Alignment for Deep Domain Adaptation},
  booktitle = {Computer Vision -- ECCV 2016 Workshops},
  year = {2016}, pages = {443--450},
  doi = {10.1007/978-3-319-49409-8_35}
}

@article{xu2026tfstsdka,
  author = {Xu, Chen-Yang and Fan, Fei-Yi and Zhao, Li-Xuan and Jin, Li-Cheng and Zhang, Yong-Hui and Meng, Qing-Hao},
  title = {Major depressive disorder detection via temporal-frequency-spatial transformer with sub-domain knowledge alignment using {EEG}},
  journal = {Neural Networks},
  year = {2026}, volume = {193}, pages = {107965},
  doi = {10.1016/j.neunet.2025.107965},
  note = {Published online 6 August 2025}
}

@article{rehman2025advances,
  title={Advances in EEG-based detection of Major Depressive Disorder using shallow and deep learning techniques: A systematic review},
  author={Rehman, Hoorain and Khan, Danish M and Amanullah, Hafsa and Kamran, Laiba and Rehman, Owais Ur and Siddiqui, Sana Taj and Masroor, Komal},
  journal={Computers in Biology and Medicine},
  volume={192},
  pages={110154},
  year={2025},
  publisher={Elsevier}
}

@article{you2025virtual,
  title={Virtual electroencephalogram acquisition: A review on electroencephalogram generative methods},
  author={You, Zhishui and Guo, Yuzhu and Zhang, Xiulei and Zhao, Yifan},
  journal={Sensors},
  volume={25},
  number={10},
  pages={3178},
  year={2025},
  publisher={MDPI}
}

@article{habashi2023generative,
  title={Generative adversarial networks in EEG analysis: an overview},
  author={Habashi, Ahmed G and Azab, Ahmed M and Eldawlatly, Seif and Aly, Gamal M},
  journal={Journal of neuroengineering and rehabilitation},
  volume={20},
  number={1},
  pages={40},
  year={2023},
  publisher={Springer}
}

@article{carrle2023generation,
  title={Generation of synthetic EEG data for training algorithms supporting the diagnosis of major depressive disorder},
  author={Carrle, Friedrich Philipp and Hollenbenders, Yasmin and Reichenbach, Alexandra},
  journal={Frontiers in Neuroscience},
  volume={17},
  pages={1219133},
  year={2023},
  publisher={Frontiers Media SA}
}

@inproceedings{wang2026sustainable,
  title={Sustainable and Responsible ECG-Based AI Diagnostics: Masked Frequency Reconstruction with Peak-Aware Transformers},
  author={Wang, Wei and Chen, Jian and Chen, Junxin and Xu, Zeling and Zou, Yuntao and HY Tong, Henry},
  booktitle={Proceedings of the ACM Web Conference 2026},
  pages={9200--9210},
  year={2026}
}

@inproceedings{fu2026towards,
  title={Towards Glaucoma Screening in Decentralized Clinics: A Dynamic Expert-Assisted Domain-Incremental Approach},
  author={Fu, Chenglong and Yao, Jian and Qian, Pengjiang and Fu, Weiwei and Wang, Chuang and Yang, Guisong},
  booktitle={Proceedings of the ACM Web Conference 2026},
  pages={9309--9317},
  year={2026}
}

@inproceedings{wang2025cbramod,
  title={Cbramod: A criss-cross brain foundation model for eeg decoding},
  author={Wang, Jiquan and Zhao, Sha and Luo, Zhiling and Zhou, Yangxuan and Jiang, Haiteng and Li, Shijian and Li, Tao and Pan, Gang},
  booktitle={International conference on learning representations},
  volume={2025},
  pages={75310--75346},
  year={2025}
}

@article{xiao2026brainomni,
  title={Brainomni: A brain foundation model for unified eeg and meg signals},
  author={Xiao, Qinfan and Cui, Ziyun and Zhang, Chi and Chen, Siqi and Wu, Wen and Thwaites, Andrew and Woolgar, Alexandra and Zhou, Bowen and Zhang, Chao},
  journal={Advances in Neural Information Processing Systems},
  volume={38},
  pages={41179--41212},
  year={2026}
}

@article{umair2025decentralized,
  title={Decentralized EEG-based detection of major depressive disorder via transformer architectures and split learning},
  author={Umair, Muhammad and Ahmad, Jawad and Alasbali, Nada and Saidani, Oumaima and Hanif, Muhammad and Khattak, Aizaz Ahmad and Khan, Muhammad Shahbaz},
  journal={Frontiers in computational neuroscience},
  volume={19},
  pages={1569828},
  year={2025},
  publisher={Frontiers Media SA}
}

@techreport{who2025mentalhealth,
  author      = {{World Health Organization}},
  title       = {World Mental Health Today: Latest Data},
  institution = {World Health Organization},
  address     = {Geneva, Switzerland},
  year        = {2025},
  isbn        = {9789240113817},
  url         = {https://www.who.int/publications/i/item/9789240113817}
}


\appendix

\section{Normative Deviation Inference Procedures}
\label{app:inference}

\subsection{Cross-electrode Spatial Adaptation}
\label{app:spatial}

We use spherical spline interpolation to map heterogeneous
EEG montages to the 19-electrode International
10--20 system~\cite{klem1999ten}.
For MODMA, source and target electrode coordinates are
obtained from the MNE GSN-HydroCel-128 and standard
10--20 templates, respectively. Coordinates are expressed
in the MNE head coordinate system, centered at
$(0,0,0.04)$ meters, and normalized to the unit sphere.

Given an EEG recording
$\mathbf{x}_{orig}\in\mathbb{R}^{C_{orig}\times T}$,
the signal at target electrode location
$\mathbf{p}_{target}$ is interpolated as
\begin{equation}
f(\mathbf{p}_{target})
=
c_0+
\sum_{j=1}^{C_{orig}}
c_j g(\|\mathbf{p}_{target}-\mathbf{p}_j\|),
\label{eq:app_spatial_interpolation}
\end{equation}
where $\mathbf{p}_j$ is the coordinate of source electrode
$j$, and $c_0$ and $c_j$ are interpolation coefficients.
For unit-sphere coordinates, the kernel is
\begin{equation}
g(r)
=
\frac{1}{4\pi}
\sum_{n=1}^{N_g}
\frac{2n+1}{[n(n+1)]^m}
P_n\left(1-\frac{r^2}{2}\right),
\label{eq:app_spatial_kernel}
\end{equation}
where $P_n$ is the degree-$n$ Legendre polynomial.
We use smoothness order $m=4$ and truncation order
$N_g=50$.
At each time sample $\tau$, the coefficients are obtained
by solving
\begin{equation}
\begin{bmatrix}
G+\lambda I & \mathbf{1}\\
\mathbf{1}^{\top} & 0
\end{bmatrix}
\begin{bmatrix}
\mathbf{c}\\
c_0
\end{bmatrix}
=
\begin{bmatrix}
\mathbf{x}_{orig}(:,\tau)\\
0
\end{bmatrix},
\label{eq:app_spatial_system}
\end{equation}
where $G_{ij}=g(\|\mathbf{p}_i-\mathbf{p}_j\|)$,
$\lambda=10^{-5}$, and $\mathbf{1}$ is the all-ones
vector. The last row enforces
$\mathbf{1}^{\top}\mathbf{c}=0$.

\subsection{S-NDD Implementation Details}
\label{app:sndd_inference_scores}

The S-NDD branch of Algorithm~\ref{alg:ndd_inference}
performs spatial adaptation, computes deviations using
two frozen references, and produces diagnostic predictions
without iterative sample generation. The implementation
samples one timestep and one noise realization per window.

\noindent\textbf{Inference-score Definitions.}
For a preprocessed EEG window
$x\in\mathbb R^{C\times L}$, the noised input and
reconstructed signal are computed as
\begin{align}
x_t
&=\sqrt{\bar\alpha_t}x
+\sqrt{1-\bar\alpha_t}\epsilon,
\label{eq:x}\\
\hat{x}_0
&=\frac{x_t-\sqrt{1-\bar\alpha_t}
\epsilon_\theta(x_t)}
{\sqrt{\bar\alpha_t}+10^{-8}},
\label{eq:supp_reconstruction}
\end{align}
where $t\sim\mathcal U\{0,\ldots,T_{\mathrm{diff}}-1\}$,
$\epsilon\sim\mathcal N(0,I)$, and
$\bar\alpha_t=\prod_{s=0}^{t}(1-\beta_s)$.
The structural and distributional deviations are
\begin{align}
D_{\mathrm{str}}(x)
&=\frac{
\|x-D_{\mathrm{vq}}(Q(E_{\mathrm{vq}}(x)))\|_F^2
}{CL},
\label{eq:supp_ndd_structure}\\
D_{\mathrm{dist}}(x)
&=\frac{\|x-\hat{x}_0\|_F^2}{CL}.
\label{eq:supp_ndd_distribution}
\end{align}
The timestep determines the noise level but is not
an explicit predictor input. Structural scoring excludes
codebook and commitment losses, whereas distributional
scoring uses reconstruction error rather than the
noise-prediction training loss.
The joint deviation score and its normalized form are
\begin{align}
S(x)
&=\alpha D_{\mathrm{str}}(x)
+(1-\alpha)D_{\mathrm{dist}}(x),
\label{eq:supp_ndd_fusion_score}\\
S'(x)
&=\operatorname{clip}\left(
\frac{S(x)-P_5}{\widetilde P_{95}-P_5},
0,1\right).
\label{eq:supp_ndd_normalization}
\end{align}
Here, $P_5$ and $P_{95}$ are estimated from all
training-partition windows, including HC and MDD.
We set $\widetilde P_{95}=P_{95}$ unless
$P_{95}-P_5<10^{-8}$, in which case
$\widetilde P_{95}=P_5+1$.
These bounds remain fixed during inference.

\begin{algorithm}[!t]
\caption{Inference pipelines of S-NDD and F-NDD}
\label{alg:ndd_inference}
\begin{algorithmic}[1]
\small
\setlength{\itemsep}{0.4pt}

\REQUIRE Formulation $m\in\{\mathrm{S},\mathrm{F}\}$;
EEG windows $\{x_{orig,i}\}_{i=1}^{N}$
($N=1$ for S-NDD);
fixed models and normalization statistics;
$\alpha,P_5,P_{95}$ for S-NDD;
optional HC calibration parameters for F-NDD.
\ENSURE Prediction
$\hat y\in\{\mathrm{MDD},\mathrm{HC}\}$.

\STATE Adapt each window:
$x_i=f_{adapt}(x_{orig,i})
\in\mathbb R^{19\times L}$.

\IF{$m=\mathrm{S}$}
    \STATE \textbf{Scalar-level inference}
    \STATE Set $x=x_1$ and reconstruct
    $\hat x=D_{vq}(Q(E_{vq}(x)))$.
    \STATE Compute $D_{str}$ using
    Eq.~\eqref{eq:supp_ndd_structure}.
    \STATE Sample
    $t\sim\mathcal U\{0,\ldots,T_{diff}-1\}$
    and $\epsilon\sim\mathcal N(0,I)$.
    \STATE Obtain $D_{dist}$ using
Eqs.~\eqref{eq:x}--\eqref{eq:supp_ndd_distribution},
with noise predictor $\epsilon_\theta(x_t)$.

    \STATE Compute $S$ and $S'$ using Eqs.~\eqref{eq:supp_ndd_fusion_score} and~\eqref{eq:supp_ndd_normalization}.
    \STATE Extract
    $F_{deep}=\mathrm{Extractor}(x)$.
    \STATE Fuse
    $F_{final}=\sigma(W_hF_{deep}+W_sS'+b)$.
    \STATE Predict
    $\hat y=h_\psi(F_{final})$.

\ELSE
    \STATE \textbf{Feature-level inference}

    \FOR{each selected F-NDD model}
        \FOR{each window $i=1,\ldots,N$}
            \STATE Extract $h_i=H(x_i)\in\mathbb R^{456}$
            and standardize using fixed HC statistics.
            \STATE Compute
            $d_{str,i}=h_i-D(Q(E(h_i)))$.
            \STATE Obtain
            $\hat h_{0,i}^{(t,k)}$ for
            $t\in\{25,125,250\}$,
            $k\in\{1,2\}$.
            \STATE Compute
            $d_{dist,i}=\frac16\sum_{t,k}
            (h_i-\hat h_{0,i}^{(t,k)})$.
        \ENDFOR

        \STATE Prepare diagnostic inputs using
        fixed source-HC scaling and
        $\operatorname{asinh}$ transforms.
        \STATE Encode all windows:
        $u_i=E_{EEG}(h_i)$,
        $v_i=P_{dev}([E_{str}(d_{str,i});
        E_{dist}(d_{dist,i})])$.
        \STATE Compute
        $a_i=\mathrm{MaskedSoftmax}_i(w^\top u_i+b)$.
        \STATE Pool
        $h_{EEG}=\sum_i a_i u_i$,
        $h_{dev}=\sum_i a_i v_i$.
        \STATE Form $F_F=[h_{EEG};h_{dev}]$
        and obtain logits from $C_F(F_F)$.
        \STATE Apply HC logit calibration
        if required by the evaluation protocol.
    \ENDFOR

    \STATE Ensemble by averaging calibrated logits
    followed by sigmoid, or averaging uncalibrated
    MDD probabilities.
    \STATE Obtain $\hat y$ using threshold $0.5$.
\ENDIF

\RETURN $\hat y$
\end{algorithmic}
\end{algorithm}

\subsection{F-NDD Implementation Details}
\label{app:fndd_inference}
\label{app:fndd_feature_extraction}

Algorithm~\ref{alg:ndd_inference} summarizes the inference
procedures of S-NDD and F-NDD. For its F-NDD branch,
this subsection details the construction of the
456-dimensional handcrafted features and the fixed
source-HC normalization used to prepare reference-model
and diagnostic inputs. The feature-space VQ-VAE and DDPM
references are trained independently on source-training
HC features and remain frozen during diagnostic inference.

\noindent\textbf{Window Processing and Spectral Estimation.}
Each EEG window is sampled at 256~Hz. Each channel is
demeaned and divided by its within-window temporal standard
deviation, with a denominator floor of $10^{-12}$.
Welch's method estimates the one-sided power spectral
density using 512-sample Hann windows, 256-sample overlap,
and a 512-point FFT. Each segment is mean-detrended,
and its periodogram is averaged with those of the
other segments.

\noindent\textbf{Spectral Features.}
We use five half-open frequency bands:
$[0.5,4)$, $[4,8)$, $[8,13)$, $[13,30)$, and
$[30,45)$~Hz. For electrode $c$, let $P_{c,b}$ be
the sum of spectral-density values in band $b$.
The log-relative power is computed as
\begin{equation}
\begin{split}
q_{c,b}
&=\frac{P_{c,b}}
{\max\!\left(\sum_{\ell=1}^{5}P_{c,\ell},10^{-12}\right)},\\
s_{c,b}
&=\log\!\left(\max(q_{c,b},10^{-8})\right),
\end{split}
\label{eq:app_fndd_spectral_features}
\end{equation}
where $\log$ denotes the natural logarithm.
The 19 electrodes contribute $19\times5=95$
spectral entries.

\noindent\textbf{Correlation Features and Ordering.}
Pearson correlations $r_{cd}$ are computed between
channel pairs within the same window, clipped to
$[-0.98,0.98]$, and transformed as
$z_{cd}=\operatorname{arctanh}(r_{cd})$.
The diagonal is then set to zero.
All $19\times19=361$ entries are retained.
For each electrode, its five spectral entries are
concatenated with its row of 19 transformed correlations.
The resulting 24-dimensional vectors are concatenated
in the fixed electrode order to obtain
$\mathbf f\in\mathbb R^{456}$, where
$456=19(5+19)=95+361$.
Symmetric duplicates and zero diagonal entries are
retained, so these are not 456 independent features.
The extraction rules have no trainable parameters.

\noindent\textbf{Source-HC Normalization.}
Here $\mathbf f$ denotes the features before
source-HC normalization. For each source-training fold,
$m_j$ and $a_j$ are the median and IQR of coordinate
$f_j$, estimated exclusively from HC windows in that
fold's training partition. The reference-model input is
\begin{equation}
h_j=\frac{f_j-m_j}{\max(a_j,0.1)}.
\label{eq:app_fndd_input_normalization}
\end{equation}
The main-text reference and residual computations use
$\mathbf h$ with this normalization understood.
Diagnostic input preparation applies
$\operatorname{asinh}(\mathbf h)$ to the baseline branch.
Signed residual coordinates are separately centered
and scaled using their source-training HC medians
and IQRs, with a scale floor of 0.05, before the
$\operatorname{asinh}$ transform and deviation encoding.
These fixed operations are included in the diagnostic
encoder input preparation.
All normalization statistics remain fixed for validation,
within-source testing, and cross-source evaluation.
Target HC used for post-hoc score calibration does not
contribute to these feature or residual statistics.

\section{Cross-Dataset Evaluation Protocols}
\label{app:transfer_protocols}

\noindent\textbf{Source Evaluation and Partitioning.}
Data splits and prediction aggregation follow the main
manuscript. Normative references are trained only on HC
examples from the training partition, whereas diagnostic
training uses labeled HC and MDD examples.
Training-derived normalization statistics, including
Eq.~\eqref{eq:supp_ndd_normalization}, remain fixed
during validation and testing.
Multiple windows from one individual are not treated
as independent subjects.

\noindent\textbf{Mumtaz2016 to MODMA.}
S-NDD is trained on Mumtaz2016 and evaluated on MODMA
without updating model weights or normalization statistics
using target data.
MODMA recordings are mapped to the 19-electrode montage
as described in Appendix~\ref{app:spatial}.
The source-trained references and diagnostic classifier
remain fixed, yielding the source-only, zero-shot
setting reported in the main manuscript.

\noindent\textbf{MODMA to Mumtaz2016.}
F-NDD is trained on MODMA and evaluated on Mumtaz2016
at the record-group level.
Each calibrated run uses seven target HC record groups,
excluded from diagnostic testing, to estimate
prediction-score calibration parameters.
Both the normative references and diagnostic network
remain fixed.
Uncalibrated and calibrated predictions are distinguished;
the latter constitute HC-assisted transfer rather than
strict source-only evaluation.
Test record groups overlap across calibration runs,
so these runs do not represent independent subject cohorts.

\noindent\textbf{Domain Adaptation Comparisons.}
The adaptation comparisons include DANN, DAN,
Deep CORAL, and TFST-SDKA in both transfer directions.
Target-data adaptation is distinguished from source-only
prediction and post-hoc HC score calibration.
In particular, adaptation restricted to target HC data
is not equivalent to UDA with unrestricted unlabeled
target data.
Differences in target-data access and prediction
aggregation must be considered when interpreting
comparisons. The S-NDD forward and F-NDD reverse
settings do not constitute a controlled comparison
of deviation granularity.

\section{Generative Baseline Implementation}
\label{sec: implementation_details}
This section details four generative
augmentation strategies developed in this
study based on DiffMDD~\shortcite{wang2024diffmdd}:
Diffusion Augmentation (Diff-Aug), hard-negative-guided synthesis (Hard Neg),
artifact cleaning (Clean), and spatial mixing (Hybrid).
Unless otherwise specified, all methods
are implemented in PyTorch and optimized
using Adam with a batch size of 32
on a NVIDIA RTX 6000 GPU.

















\begin{table}[t]
  \caption{Stage-wise training and sampling configurations
  of Diff-Aug. lr and wd denote the learning rate and
  weight decay, respectively.}

\vspace{-1em}
  \label{tab:diff_aug_settings}
  \centering
  \begingroup
  \footnotesize
  \setlength{\tabcolsep}{3pt}
  \renewcommand{\arraystretch}{1.02}
  \begin{tabularx}{\columnwidth}{@{}
    >{\raggedright\arraybackslash}p{0.23\columnwidth}
    >{\raggedright\arraybackslash}X@{}}
    \toprule
    \textbf{Stage} & \textbf{Configuration} \\
    \midrule
    Condition classifier &
    1D-CNN--Transformer; Adam, lr $=10^{-3}$,
    wd $=10^{-4}$; dropout $=0.5$, patience $=8$. \\
    \midrule
    Diffusion training &
    1D-CNN--Transformer with FiLM;
    $T=1000$, $\beta\in[10^{-4},0.02]$;
    MSE, lr $=10^{-3}$, 50 epochs, patience $=8$. \\
    \midrule
    Data synthesis &
    570 anchors, $k=3$ per anchor;
    100-step DDIM, guidance $=1.0$. \\
    \midrule
    Downstream training &
    1D-CNN--Transformer; lr $=10^{-4}$,
    100 epochs; $\delta=0.1$, patience $=15$. \\
    \bottomrule
  \end{tabularx}
  \endgroup
  
\end{table}

\begin{table}[t]
\caption{Key configurations of Hard Neg. Settings not
listed here follow those of Diff-Aug.}
\vspace{-1em}
  \label{tab:hard_negative_settings}
  \centering
  \begingroup
  \footnotesize
  \setlength{\tabcolsep}{3pt}
  \renewcommand{\arraystretch}{1.02}
  \begin{tabularx}{\columnwidth}{@{}
    >{\raggedright\arraybackslash}p{0.23\columnwidth}
    >{\raggedright\arraybackslash}X@{}}
    \toprule
    \textbf{Stage} & \textbf{Configuration} \\
    \midrule
    Anchor selection &
    $s_i=0.7\widetilde D_{\mathrm{str}}
    +0.3\widetilde D_{\mathrm{dist}}$;
    normalization: 5th--95th percentiles;
    hard range: 70th--90th percentiles;
    exclude the top 10\%. \\
    \midrule
    Data synthesis &
    Diff-Aug predictor; 100-step DDIM;
    classifier guidance $=2.5$;
    five pseudo-segments per anchor. \\
    \midrule
    Downstream training &
    $H_i\in\mathbb{R}^{L\times128}$,
    $W_{\mathrm{att}}\in\mathbb{R}^{128\times1}$;
    attention: $\operatorname{Softmax}_{L}
    (s_iH_iW_{\mathrm{att}})$;
    optimization as in Diff-Aug. \\
    \bottomrule
  \end{tabularx}
  \endgroup
  
\end{table}

\noindent\textbf{(i) Diff-Aug.}
Diff-Aug evaluates synthetic data augmentation
as an alternative to normative deviation
modeling, retaining the same downstream
1D-CNN--Transformer classifier.
Following DiffMDD, the pipeline consists of
three stages. First, a 1D-CNN--Transformer
classifier is pre-trained to extract
class-aware condition embeddings.
Next, a conditional diffusion model employs
a 1D-CNN--Transformer noise predictor,
incorporating these embeddings through
Feature-wise Linear Modulation (FiLM).
Finally, synthetic EEG segments are generated
from randomly selected training anchors
using classifier-guided DDIM sampling
and combined with the original training
data to train the classifier from scratch.
Implementation settings are summarized
in Table~\ref{tab:diff_aug_settings}.

\noindent\textbf{(ii) Hard Neg.}
\label{subsec:hard_negative}
Hard Neg selects training anchors based on
normative deviations for conditional EEG
generation and deviation-conditioned classification.
For each sample $x_i$, the joint score is
$s_i=0.7\widetilde{D}_{\mathrm{str}}(x_i)
+0.3\widetilde{D}_{\mathrm{dist}}(x_i)$,
where the structural and distributional deviations
are normalized using the 5th and 95th percentiles
of the training distribution.
Samples within the 70th--90th score percentiles
are selected as hard anchors, excluding the
highest 10\%. Following Diff-Aug, conditional
generation uses 100-step DDIM sampling with
classifier guidance of 2.5, producing five
pseudo-segments per anchor.
The generated samples are combined with the
original training data for classifier training.
Given sequence features
$H_i\in\mathbb{R}^{L\times128}$ and a learnable
projection $W_{\mathrm{att}}\in\mathbb{R}^{128\times1}$,
the sample-specific scalar $s_i$ modulates
attention as
$A_i=\operatorname{Softmax}_{L}
(s_iH_iW_{\mathrm{att}})$,
where $L$ is the sequence length and
$A_i\in\mathbb{R}^{L\times1}$.
The attended representation is
$h_i^{\mathrm{att}}=H_i^\top A_i
\in\mathbb{R}^{128}$.
Other optimization settings follow Diff-Aug,
with configurations summarized in
Table~\ref{tab:hard_negative_settings}.

\noindent\textbf{(iii) Clean.}
Clean follows the Diff-Aug generation and
classification pipeline, applying a fourth-order
Butterworth low-pass filter with a 50 Hz cutoff
to each generated segment:
$\widetilde{x}_{\mathrm{gen}}
=\mathcal{B}_{50}(x_{\mathrm{gen}})$.
The filter is implemented using
\texttt{scipy.signal.filtfilt} for zero-phase
forward--backward filtering.
The filtered segments are used for 
classifier training, with settings summarized
in Table~\ref{tab:clean_settings}.

















\begin{table}[t]
\caption{Key configurations of Clean. Settings not
listed here follow those of Diff-Aug.}
\vspace{-1em}
  \label{tab:clean_settings}
  \centering
  \begingroup
  \footnotesize
  \setlength{\tabcolsep}{3pt}
  \renewcommand{\arraystretch}{1.02}
  \begin{tabularx}{\columnwidth}{@{}
    >{\raggedright\arraybackslash}p{0.23\columnwidth}
    >{\raggedright\arraybackslash}X@{}}
    \toprule
    \textbf{Stage} & \textbf{Configuration} \\
    \midrule
    Data synthesis &
    Diff-Aug predictor; 100-step DDIM;
    classifier guidance $=1.0$. \\
    \midrule
    Artifact cleaning &
    Fourth-order Butterworth low-pass filter,
    cutoff $=50$ Hz; zero-phase forward--backward
    filtering via \texttt{scipy.signal.filtfilt}. \\
    \midrule
    Downstream training &
    Original and filtered pseudo-segments;
    classifier and optimization as in Diff-Aug. \\
    \bottomrule
  \end{tabularx}
  \endgroup
  
\end{table}





\noindent\textbf{(iv) Hybrid.}
Hybrid introduces spatial mixing to increase cross-electrode diversity
in generated EEG segments. The generated segments are used for
downstream supervised training under the reported evaluation splits.

\section{Computational Efficiency Analysis}
\label{efficiency_analysis}

Table~\ref{tab:efficiency} summarizes the computational
cost and inference latency of different methods.
All latency measurements are conducted on an
NVIDIA RTX 6000 GPU.
Unlike generative augmentation methods, S-NDD
and F-NDD estimate normative deviations without
iterative sample synthesis.
S-NDD requires 1.70M parameters and 1.42 GFLOPs,
while the native single-model F-NDD contains
0.457M parameters and requires 0.00405 GFLOPs
of profiled neural operations.
The inference latency is approximately 2.9 ms
for S-NDD and 16.5 ms for F-NDD.
F-NDD latency includes feature extraction and
data transfers, while its FLOPs exclude
CPU feature processing. In practice, the fixed cost
of S-NDD stems from one VQ-VAE reconstruction pass,
one denoising score evaluation, and the diagnostic
classifier. In contrast, DiffMDD relies on 100-step
DDIM synthesis, which dominates its 101.69 GFLOPs and
approximately 94.5 ms latency; S-NDD thus reduces the
FLOPs by about $72\times$ and the latency by more than
$30\times$ with comparable parameters (1.70M versus
1.32M).

\begin{table}[t]
  \centering
  \caption{Computational cost and inference latency
  measured on an NVIDIA RTX 6000 GPU.}
  \vspace{-1em}
  \label{tab:efficiency}

  \begingroup
  \scriptsize
  \setlength{\tabcolsep}{3pt}
  \renewcommand{\arraystretch}{1.12}

  \resizebox{\columnwidth}{!}{%
  \begin{tabular}{@{}lcccc@{}}
    \toprule
    \textbf{Method}
    & \textbf{Synthesis}
    & \textbf{Params (M)}
    & \textbf{FLOPs (G)}
    & \textbf{Latency (ms)} \\
    \midrule

    1D-CNN-Trans
    & $\times$
    & 0.59
    & 0.14
    & $\sim$0.6 \\

    DiffMDD$^{*}$
    & $\checkmark$
    & 1.32
    & 101.69
    & $\sim$94.5 \\

    \midrule

    S-NDD (Ours)$^{\dagger}$
    & $\times$
    & 1.70
    & 1.42
    & $\sim$2.9 \\

    F-NDD (Ours)$^{\ddagger}$
    & $\times$
    & 0.457
    & 0.00405
    & $\sim$16.5 \\

    \bottomrule
  \end{tabular}%
  }

  \par\vspace{2pt}
  \begin{minipage}{\columnwidth}
    \scriptsize
    $^{*}$Includes 100-step DDIM synthesis.
    $^{\dagger}$Includes one denoising score evaluation.
    $^{\ddagger}$Native single-model F-NDD;
    latency includes feature processing and
    data transfers, while FLOPs cover
    profiled neural operations.
  \end{minipage}

  \endgroup
\end{table}

\end{document}